\documentclass{article}

\PassOptionsToPackage{numbers, compress}{natbib}

 \usepackage[preprint]{skill-nb}

\usepackage[utf8]{inputenc} 
\usepackage[T1]{fontenc}    
\usepackage{hyperref}       
\hypersetup{hidelinks}
\usepackage{url}            
\usepackage{booktabs}       
\usepackage{amsfonts}       
\usepackage{nicefrac}       
\usepackage{microtype}      
\usepackage{xcolor}         
\usepackage{graphicx}
\usepackage{wrapfig}
\usepackage{multirow}
\usepackage{tabularx}
\usepackage{tablefootnote}
\usepackage{float}
\usepackage{paralist}
\usepackage{enumitem}
\usepackage{amsmath}
\usepackage{amssymb}
\usepackage{algorithm}
\usepackage{algpseudocode}
\usepackage{tikz}
\usepackage{pgfplots}
\pgfplotsset{compat=1.18}
\usepgfplotslibrary{groupplots}
\usetikzlibrary{shapes.geometric, arrows.meta, positioning, fit, backgrounds, calc}

\title{\textsf{SKILL.nb}: Selective Formalization and Gated~Execution for Durable Agent Workflows}
\author{%
  Amine El Hattami\textsuperscript{1,2,3},
  Nicolas Chapados\textsuperscript{1},
  Christopher Pal\textsuperscript{1,2,3,4}\\
  \textsuperscript{1}ServiceNow Research,
  \textsuperscript{2}Mila,
  \textsuperscript{3}Polytechnique Montr\'eal,
  \textsuperscript{4}Canada CIFAR AI Chair
}

\begin{document}

\maketitle

\begin{abstract}
AI agents increasingly convert past experience into reusable artifacts such as code, workflows, and procedural memories.
Reuse improves efficiency but creates a lifecycle reliability problem: artifacts that succeed once may fail under environment drift, underspecified tasks, or changing task distributions, especially in web automation.
We introduce \textsf{SKILL.nb}, a framework for governing reusable agent workflows through evidence-calibrated lifecycle policies.
Its key mechanism is \emph{selective formalization}: execution evidence decides which workflow steps should become executable code, which should remain natural-language-guided, and when those choices should be revised.
\textsf{SKILL.nb} stores workflows as auditable, versioned notebooks that interleave natural-language guidance, multi-language executable cells, validation gates, fallback paths, and multimodal evidence such as outputs, screenshots, and error traces.
At runtime, \textsf{SKILL.nb} performs \emph{gate-conditioned execution}: unlike all-or-nothing scripts, each step can execute code when its gates validate, or fall back locally to an NL procedure or step intent when drift invalidates the executable realization.
Cell-level records of attempted realizations, gate outcomes, outputs, screenshots, and fallbacks make both workflow updates and executions auditable.
On WebArena-Verified, \textsf{SKILL.nb} achieves 53.7\% single-round success, improving over the strongest baseline by 3.9 percentage points.
Across three re-executions, it retains 91.7\% of initially successful tasks, 15.5 points above the next best method.
Under bounded repair, it recovers 72.9\% of subsequent failures while limiting post-repair regressions to 4.2\%, compared with 15.0--17.0\% regression rates for persistent baselines.
It also leads the compared methods on Mind2Web cross-website and cross-domain splits.
In a realistic GitLab migration test, \textsf{SKILL.nb} preserves performance when reusing frozen state learned on GitLab~15.7, with frozen-versus-fresh target-version gaps of only \(-1.7\) points on GitLab~16.11 and \(+0.6\) points on GitLab~18.9; the least-degraded persistent baseline drops by \(10.6\)--\(11.1\) points.
These results identify lifecycle governance and gate-conditioned execution as reliability axes beyond one-shot task success.
Code, data, and evaluation scripts are available at \url{https://github.com/Am1n3e/skill-nb.git}.
\end{abstract}

\section{Introduction}

AI agents increasingly produce and rely on durable external artifacts such as code, workflows, and procedural memories~\citep{liu2024large,wuOSCopilotGeneralistComputer2024,wang2024agentworkflowmemory,ouyang2025reasoningbank}.
As these artifacts are reused, the central question shifts from one-shot task completion to whether the artifact remains effective as conditions change, such as when the target site for a web agent changes its UI or data layout~\citep{levy_st-webagentbench_2025}.
Recent memory and workflow systems treat agent experience as reusable~\citep{zhengSynapseTrajectoryasExemplarPrompting2023a,zhao2024expel,fang2025memp,wang2024agentworkflowmemory,ouyang2025reasoningbank}, but repeated use introduces software-maintenance lifecycle concerns~\citep{rahman2023maintenance}: artifacts need versioning, validation, repair, regression control, and retirement when assumptions fail.
Existing memory-centric systems often reuse experience as prompt context, while workflow and artifact systems provide execution, validation, versioning, or state-management primitives~\citep{wang2024agentworkflowmemory,ouyang2025reasoningbank,reuseit2025,alas2025,agentgit2025,atomix2026}.
These mechanisms are typically studied separately rather than as a joint lifecycle policy.
For sustained automation, the question is not merely how to recall prior experience, but how accumulated execution evidence should govern reusable workflows: when to promote a candidate, when to formalize a step, when to repair or demote brittle realizations, and when to retire a workflow whose assumptions no longer hold.

We study this lifecycle problem in web environments because they provide a high-drift setting for evaluating learned artifacts.
Site changes are outside the automation system's control: unlike desktop applications that can often be version-pinned, or APIs that typically expose explicit contracts and announce breaking changes, web interfaces can drift without preserving automation anchors such as selectors or layouts.
While APIs are often the preferred automation interface, many real workflows still depend on UI-level interaction~\citep{zhangYouOnlyLook2024}.
Automating web workflows therefore forces a formalization choice: which steps should be hardened into code, and which should remain natural-language-guided (NL-guided) when interface drift would make repeated repair too costly.
As a limiting case, mathematical autoformalization illustrates the fully formal endpoint: translating informal arguments into proof-assistant artifacts such as \emph{Lean} yields terms checked by a well-defined logical kernel~\citep{demoura2021lean4,wu2022autoformalization,jiang2023draft}.
Durable agent workflows occupy an intermediate point on this spectrum.
Leaving steps NL-guided preserves flexibility but weakens artifact-level control, while hardening them into code makes reuse more controlled but can become brittle under drift.
Their validity therefore depends both on a task-level procedure and on an environment-dependent realization whose assumptions can break.
For durable workflows, formalization is therefore lifecycle-governed.
The system must decide when to create or retire a workflow, which steps remain NL procedures versus executable code, and when execution evidence warrants revising those choices as interfaces drift.

We introduce \textsf{SKILL.nb}, a framework for learning evidence-based lifecycle policies over reusable agent workflows.
Its central mechanism is \emph{selective formalization}: using execution evidence to decide which steps remain NL-guided, which become executable code, and when those choices should be revised.
\textsf{SKILL.nb} couples this with \emph{gate-conditioned execution}: runtime gates decide whether to execute code or fall back to an NL procedure.
Each workflow is represented as an auditable, versioned notebook that interleaves NL guidance, executable code, validation gates, fallback paths, and cell-level evidence.
Appendix~\ref{app:artifact_example} visualizes how a provisional task notebook is promoted into a released workflow.
Offline lifecycle learning aggregates traces into evidence counts for workflow creation and step formalization, and into accepted-repair signals, including repair counts and token-weighted repair burden as a proxy for instability in step demotion and workflow retirement.
Thresholds are calibrated by replaying candidate lifecycle policies on historical workflows, selecting low-maintenance policies whose validation-failure rate stays within budget.
This calibration uses accumulated execution evidence rather than hidden evaluator labels, and remains replay-relative rather than a guarantee under future interface shifts.
Because web drift makes hardened code useful but brittle, the representation must
preserve execution evidence, not just instructions. Otherwise, maintenance cannot
tell whether code was merely suggested or actually ran, which gates validated it,
what evidence it produced, or when execution fell back. This raises a natural
question: why not use a conventional \texttt{SKILL.md} file with instructions and
code snippets? Markdown can describe code, but it does not by itself record
execution boundaries, gate outcomes, fallbacks, or cell-local evidence.
\textsf{SKILL.nb} makes code execution first-class and auditable by storing these
objects as versioned notebooks. Appendix~\ref{app:skill_md_vs_nb} compares the
two representations along four axes: executable workflow state, validation
feedback, evidence retention, and failure localization.

Empirically, \textsf{SKILL.nb} improves one-shot task performance and controlled repeated-use reliability.
On WebArena-Verified~\citep{elhattami2025webarenaverified}, it achieves the highest single-round success rate among compared methods---CodeAct~\citep{wang2024executablecodeactionselicit}, AWM\textsubscript{online}~\citep{wang2024agentworkflowmemory}, and ReasoningBank~\citep{ouyang2025reasoningbank}---at 53.7\%, outperforming the next-best method by 3.9 percentage points (paired McNemar, $p=0.029$).
In controlled repeated execution, \textsf{SKILL.nb} retains 91.7\% of initially successful WebArena-Verified tasks across three re-executions, 15.5 percentage points above the next best method.
Under bounded repair, it recovers 72.9\% of subsequent failures while limiting post-repair regressions to 4.2\%, compared with 15.0--17.0\% for persistent baselines under their native update paths.
As a secondary transfer evaluation on Mind2Web~\citep{dengMind2WebGeneralistAgent2023a}, \textsf{SKILL.nb} leads the compared methods on all four metrics in both the cross-website and cross-domain splits, with step success reaching 38.1\% and 39.7\%, respectively.
In a controlled GitLab migration test, \textsf{SKILL.nb} preserves target-version performance when reusing frozen state learned on GitLab~15.7, while persistent-memory baselines degrade by 10.6--14.4 points.
These results show that governed workflow artifacts improve task completion, reuse, repair, and regression control under those evaluated protocols.

Our key contributions are summarized as follows:
\begin{itemize}[leftmargin=*, itemsep=0pt, topsep=1pt]
\item We formulate durable web-agent automation as \emph{selective formalization} and lifecycle governance for reusable, executable artifacts that must be formalized, validated, repaired, and retired as assumptions fail.
	\item We present \textsf{SKILL.nb}, a notebook-native framework that couples selective formalization with gate-conditioned execution, enabling versioned workflows that mix NL guidance, executable cells, validation gates, fallback, and auditable multimodal execution evidence.
	\item We evaluate \textsf{SKILL.nb} on WebArena-Verified and Mind2Web, showing the highest single-round performance among compared methods and improved retention, repair recovery, and regression control in repeated WebArena-Verified execution, and include a controlled GitLab version-drift protocol comparing fresh runs with old-state reuse after application migration.
\end{itemize}

\section{Related Work}

\noindent\textbf{Agent Experience Memory.} Memory is an essential module for agents~\citep{zhang2024memorymechanism}, with representations ranging from virtual paging~\citep{packer2023memgpt} and structured graphs~\citep{chhikara2025mem0,xu2025agenticmem} to hierarchical working memory and consolidation mechanisms~\citep{hu2025hiagent,tan2025reflectivememory,zhongMemoryBankEnhancingLarge2024}. A complementary line of work focuses on reusing past experience for future tasks. For instance, Synapse retrieves raw trajectories as in-context exemplars~\citep{zhengSynapseTrajectoryasExemplarPrompting2023a}, while ExPeL and MemP distill experience into procedural insights and memory~\citep{zhao2024expel,fang2025memp}. Extending this, Agent Workflow Memory (AWM) and ReasoningBank abstract execution traces into reusable workflows and reasoning strategies~\citep{wang2024agentworkflowmemory,ouyang2025reasoningbank}. Crucially, across all these systems, retrieved experience remains \emph{advisory} prompt context: it guides generation but is not natively executed, verified, or version-controlled. \textsf{SKILL.nb} bridges this gap by operationalizing memory into governed, executable workflow artifacts validated by deterministic acceptance checks rather than stochastic LLM judgments.

\noindent\textbf{Self-Evolving Agents.} Continuous adaptation is a core desideratum for autonomous agents~\citep{gao2025selfevolving,liang2024selfevolving}. To achieve cross-task evolution, systems maintain refined causal abstractions (CLIN~\citep{majumder2023clin}), construct training curricula~\citep{su2025learnbyinteract}, or consolidate transferable reasoning strategies (ICE~\citep{qian2024ice}, ChemAgent~\citep{tang2025chemagent}, Contextual Replay~\citep{liu2025contextualreplay}). More closely related to our approach are Voyager and TroVE, which construct open-ended skill libraries of executable code through exploration~\citep{wangVoyagerOpenEndedEmbodied2023,wang2024trove}. However, because these methods typically append distilled lessons or skills to memory without strict lifecycle gating, a single erroneous update can silently corrupt an agent's future behavior. \textsf{SKILL.nb} mitigates this risk by enforcing governed evolution: candidate updates enter the repository only after offline verification and deterministic gate checks, enabling safe rollbacks if regressions occur.

\noindent\textbf{Durable Agent Artifacts.} As agents produce durable outputs, governing their creation, validation, and maintenance becomes critical. Systems typically manage this evolution through versioned execution logs and localized repair (ALAS~\citep{alas2025}), git-like state checkpointing (AgentGit~\citep{agentgit2025}), or iterative notebook refinement~\citep{you2025datawiseagentnotebookcentricllmagent,elhashemy2025bridgingprototypeproductiongapmultiagent}. At the runtime level, Atomix enforces strict transactional semantics for tool calls~\citep{atomix2026}, while ReUseIt synthesizes guarded workflows from repeated attempts~\citep{reuseit2025}. However, ReUseIt validates these guards via stochastic LLM screenshot analysis, and other complementary systems rely heavily on human-in-the-loop oversight during failures~\citep{agentbay2025,huq-etal-2025-cowpilot} or external skill induction~\citep{tao-etal-2025-treerag,zheng2025skillweaverwebagentsselfimprove}. \textsf{SKILL.nb} differentiates itself from these approaches by governing artifact lifecycles entirely offline, replacing stochastic runtime validation and human oversight with deterministic gate checks and trace-replay calibration.

\noindent
Taken together, prior work supplies isolated primitives for memory, workflow induction, and artifact governance. \textsf{SKILL.nb} integrates these strands into a unified lifecycle, complementing memory-reuse frameworks, skill-induction systems, and artifact managers.

\section{\textsf{SKILL.nb}}\label{sec:skillnb}

We present \textsf{SKILL.nb}, an online--offline framework for governing reusable workflow artifacts with an RLVR-inspired preference for execution-grounded evidence~\citep{lambert2024tulu,guo2025deepseek}. 
Its core mechanism is \emph{selective formalization}: deciding when steps remain NL-guided, become executable, are demoted after instability, or contribute to workflow retirement. 
We describe the artifact representation, lifecycle policy, adaptive thresholds, and runtime loop; complete algorithms are in Appendix~\ref{app:runtime_system}.

\subsection{Versioned Workflow Artifacts}\label{ssec:workflow_artifacts}

\textsf{SKILL.nb} stores each reusable procedure as a versioned artifact in a repository $\mathcal{K}$, supporting retrieval, updates, and rollback. 
A workflow version $\mathcal{W}_v$, $v\in\{1,2,\ldots\}$, and its steps are
\[
\mathcal{W}_{v}=\langle I,X,S,M^W\rangle,\qquad
s_i=\langle I_i,P_i,C_i,\Gamma_i,M_i^S\rangle,\qquad
\Gamma_i=(\gamma_{i,\mathrm{pre}},\gamma_{i,\mathrm{post}}).
\]
Here $I$ is the workflow intent, $X$ the input schema and validation rules, $S=(s_1,\ldots,s_n)$ the ordered steps, and $M^W$ workflow metadata. 
Each step stores a local intent $I_i$, an NL procedure $P_i$, an optional executable realization $C_i$, executable pre/post gates $\Gamma_i$, and step metadata $M_i^S$.

Verification gates are predicates over environment-observable states and do not access benchmark evaluators or hidden success labels. 
Metadata supports retrieval at workflow and step level: $M^W$ indexes similar task flows, while $M_i^S$ indexes analogous or specialized steps. 
In our implementation, artifacts are Jupyter notebooks\footnote{\url{https://jupyter.org/}} interleaving these components. 
Appendix~\ref{app:artifact_example} shows an example.

\subsection{Selective Formalization as a Lifecycle Policy}\label{ssec:selective_formalization}

Selective formalization governs both workflow maturity and step representation. 
For workflow version $\mathcal{W}_v$, let
\[
y(\mathcal{W}_v)\in\{\texttt{provisional},\texttt{released},\texttt{retired}\}
\]
denote whether it is under validation, available for retrieval and execution, or removed from active retrieval while retained for rollback and analysis. 
For each step $s_i$, let $z_i\in\{0,1\}$ indicate whether a validated executable realization $C_i$ is available. 
If $z_i=0$, execution falls back to $P_i$ or, when necessary, the bare intent $I_i$; $z=(z_1,\ldots,z_n)$ is the workflow formalization pattern.

Ideally, a lifecycle controller would choose $(y,z)$ to trade offline maintenance cost against repeated runtime cost:
\[
\begin{aligned}
(y^\star,z^\star)
&=\arg\min_{y,z}\;
C_{\mathrm{maint}}(y,z\mid\mathcal{W}_v)
+ C_{\mathrm{run}}(y,z\mid\mathcal{W}_v) \\
&\text{s.t.}\quad
J_{\mathrm{perf}}(y,z\mid\mathcal{W}_v)
\le J_{\mathrm{perf}}^{\mathrm{ref}}+\epsilon .
\end{aligned}
\]
Here $C_{\mathrm{maint}}$ is cumulative offline maintenance inference cost, including distillation, validation, repair, promotion, demotion, and retirement review. 
We approximate each maintenance event $e$ by token cost $c(e)=\mathrm{tok}_{\mathrm{in}}(e)+\mathrm{tok}_{\mathrm{out}}(e)$. 
The runtime term $C_{\mathrm{run}}$ includes online inference, fallbacks, retries, and browser/tool actions. 
The loss $J_{\mathrm{perf}}$ is estimated from offline replay validation and captures downstream failure, regressions on previously passing traces, or step-level degradation relative to $J_{\mathrm{perf}}^{\mathrm{ref}}$, with tolerance $\epsilon\ge0$. 
Runtime gates do not access $J_{\mathrm{perf}}$ or hidden benchmark labels.

The full objective is not directly solved, since lifecycle actions change future repository state, trace distributions, and maintenance opportunities. 
\textsf{SKILL.nb} instead uses a restricted threshold policy $\pi_\theta$, with
$\theta=(\tau_{\mathrm{create}},\tau_{\mathrm{form}},\tau_{\mathrm{demote}},\tau_{\mathrm{retire}})$, that emits create, form, demote, or retire actions from logged evidence. 
Create, form, and demote use count thresholds; retirement uses a normalized repair burden in $[0,1]$. 
Threshold crossings trigger maintenance review, and repository state changes only after the corresponding artifact update passes validation.

The thresholds use two evidence types. 
\emph{Trace-support evidence} governs workflow creation and step formalization. 
For request $q$, let $\mathcal{C}(q)$ be the cluster of similar prior traces available before the lifecycle decision. 
The workflow-level count $|\mathcal{C}(q)|$ is compared with $\tau_{\mathrm{create}}$. 
For each step $s_i$, maintenance aligns traces in $\mathcal{C}(q)$ to the workflow sequence and counts validated execution segments supporting that step, giving $n_i^{\mathrm{evidence}}$, which is compared with $\tau_{\mathrm{form}}$.

\emph{Repair evidence} governs demotion and retirement. 
Let $m_i^{\mathrm{repair}}$ be the number of accepted repairs affecting step $s_i$ in version $\mathcal{W}_v$; only repairs passing maintenance validation are counted. 
If $m_i^{\mathrm{repair}}\ge\tau_{\mathrm{demote}}$, the executable realization for $s_i$ is demoted to NL-guided execution. 
For retirement, repair count alone is too coarse, so we use a token-weighted burden. 
Let $\mathcal{R}_i(\mathcal{W}_v)$ be accepted repairs affecting step slot $i$ along the lineage ending at $\mathcal{W}_v$, let $\mathcal{R}_{\mathrm{cal}}$ be accepted repairs in calibration logs, and set $c_{\mathrm{ref}}=\max_{e\in\mathcal{R}_{\mathrm{cal}}}c(e)$. 
If $\mathcal{R}_{\mathrm{cal}}$ is empty, automatic repair-burden retirement is deferred; otherwise,
\[
\rho_{\mathrm{repair}}(\mathcal{W}_v)
=
\frac{1}{|S|}
\sum_{i=1}^{|S|}
\min\!\left(
1,
\frac{\sum_{e \in \mathcal{R}_i(\mathcal{W}_v)} c(e)}
     {c_{\mathrm{ref}}}
\right)
\in [0,1].
\]
This averages capped per-step repair burden, preventing a single repeatedly repaired step from dominating the retirement signal.

\subsection{Adaptive Thresholds via Group Specialization}
\label{ssec:adaptive_thresholds}

A single global $\theta$ is often too aggressive for sparse domains and too conservative for well-observed ones. 
\textsf{SKILL.nb} therefore treats groups as threshold-sharing units: workflow decisions (\texttt{create}, \texttt{retire}) use workflow groups $g^W\in\mathcal{G}^W$, while step decisions (\texttt{form}, \texttt{demote}) use step groups $g^S\in\mathcal{G}^S$. 
Groups are canonicalized from artifact metadata such as site family, task type, action type, and interface properties, excluding benchmark task IDs and hidden labels. 
When a statement applies to either workflow or step groups, we write $g\in\mathcal{G}_d$, where $d$ is the lifecycle decision.

Let $\mathcal{L}_{\mathrm{thr}}$ be historical execution logs, separate from final evaluation tasks. 
A threshold-estimation case $j$ for decision $d$ contains the signal $u_j$, candidate lifecycle action, attributable maintenance token cost, and offline validation outcome; cases missing any of these fields are excluded. 
Let $\mathcal{D}_{g,d}\subseteq\mathcal{L}_{\mathrm{thr}}$ be the usable cases for group $g$ and decision $d$, with $n_{g,d}=|\mathcal{D}_{g,d}|$. 
The signal is $|\mathcal{C}(q)|$ for creation, $n_i^{\mathrm{evidence}}$ for formalization, $m_i^{\mathrm{repair}}$ for demotion, and $\rho_{\mathrm{repair}}(\mathcal{W}_v)$ for retirement. 
Let $\mathcal{T}_d$ be the sorted unique signal values in the logs; threshold $\tau$ admits case $j$ when $u_j\ge\tau$.

For candidate $\tau$, $\hat{C}^{(d)}_{\mathrm{maint}}(g,\tau)$ is the replay-estimated maintenance token cost for decision $d$ on $\mathcal{D}_{g,d}$, omitting costs common to all thresholds. 
This optimizes maintenance compute rather than full $C_{\mathrm{total}}$: runtime effects of suppressed actions are not reliably counterfactually observed; runtime efficiency, recovery, and regression are instead controlled through validation filters and measured downstream.

The validation-violation rate estimates how often $\tau$ would admit an unsafe lifecycle action:
\[
\hat{V}^{(d)}(g,\tau)=k_{g,d}(\tau)/n_{g,d},\qquad n_{g,d}>0,
\]
where $k_{g,d}(\tau)$ counts admitted cases whose offline validation loss exceeds $J_{\mathrm{perf}}^{\mathrm{ref}}+\epsilon$. 
If $n_{g,d}=0$, no group-specific violation rate is estimated. 
To avoid treating small groups as safe because they have few observed violations, a threshold is feasible only when the one-sided Wilson upper confidence bound is below the decision-specific budget $V_{\max}^{(d)}$:
\[
\mathcal{F}_{g,d}
=
\left\{
\tau\in\mathcal{T}_d:
\mathrm{WilsonUCB}_{1-\alpha}(k_{g,d}(\tau),n_{g,d})
\le V_{\max}^{(d)}
\right\}.
\]
The pooled feasible set $\mathcal{F}^{\mathrm{pool}}_d$ is computed analogously after pooling cases across groups and supplies the default for unseen or sparse groups. 
This replay filter uses only logged cases with observed outcomes; it does not simulate induced changes to later repository contents, traces, or repairs. 
Because thresholds are selected after sweeping many candidates, the Wilson bound is a conservative replay filter rather than a uniform post-selection or deployment-time safety guarantee.

If $\mathcal{F}^{\mathrm{pool}}_d\neq\emptyset$, the pooled threshold is
\[
\hat{\tau}^{\mathrm{pool}}_d
=
\arg\min_{\tau\in\mathcal{F}^{\mathrm{pool}}_d}
\hat{C}^{(d)}_{\mathrm{maint}}(\mathrm{pool},\tau).
\]
If $\mathcal{F}^{\mathrm{pool}}_d=\emptyset$, automatic thresholding for decision $d$ is deferred to maintenance review. 
For a group with $n_{g,d}>0$ and $\mathcal{F}_{g,d}\neq\emptyset$,
\[
\hat{\tau}^{g}_d
=
\arg\min_{\tau\in\mathcal{F}_{g,d}}
\hat{C}^{(d)}_{\mathrm{maint}}(g,\tau).
\]
Deployment shrinks small-group estimates toward the pooled threshold:
\[
\tau_d^g =
\begin{cases}
\Pi_{\mathcal{F}_{g,d}}\!\left[
\omega_{g,d}\hat{\tau}_d^g
+(1-\omega_{g,d})\hat{\tau}_d^{\mathrm{pool}}
\right],
& n_{g,d}>0,\;\mathcal{F}_{g,d}\ne\emptyset,\;
  \mathcal{F}^{\mathrm{pool}}_d\ne\emptyset,\\[5pt]
\Pi_{\mathcal{F}^{\mathrm{pool}}_d}\!\left[
\hat{\tau}_d^{\mathrm{pool}}
\right],
& n_{g,d}=0,\;\mathcal{F}^{\mathrm{pool}}_d\ne\emptyset,\\[5pt]
\operatorname{defer}_d,
& \text{otherwise}.
\end{cases}
\]
Here $\omega_{g,d}=n_{g,d}/(n_{g,d}+n_0)$ and $\Pi_{\mathcal{F}}$ projects to the nearest replay-supported feasible value. 
The value $\operatorname{defer}_d$ means no automatic thresholded action is taken, and the case is routed to maintenance review. 
At runtime, $\pi_\theta$ compares each lifecycle signal with the deployed group threshold; crossings trigger review, not direct writes to $\mathcal{K}$. 
Appendix~\ref{app:adaptive_thresholds} gives the Wilson formula, tie-breaking, empty-set handling, and full estimation procedure.

\paragraph{Threshold calibration versus end-to-end RL.}
Recent RLVR methods update model parameters using verifiable task rewards~\citep{lambert2024tulu,guo2025deepseek}. 
\textsf{SKILL.nb} adopts the same preference for externally checkable outcomes, but applies it to durable workflow artifacts rather than model weights. 
In web automation, failures arise from exogenous interface drift---changed DOM structure, selectors, or page flow~\citep{choudhary2011water,leotta2014visual,leotta2016robula,stocco2018vista}---so the failing object is often the artifact, not the base model. 
\textsf{SKILL.nb} therefore updates versioned artifacts while keeping the base LLM fixed during runtime and offline maintenance.

Lifecycle decisions are sparse governance actions over accumulated evidence, not token-level policy updates. 
Release, formalization, demotion, and retirement depend on auditable signals such as trace support and accepted repair counts, while replay-estimated violation rates determine feasible thresholds. 
Thresholds expose these tradeoffs, support conservative replay filtering, and remain inspectable during review. 
Online RL would require deployment-sensitive exploration over durable artifacts, while richer offline controllers would require full-trajectory counterfactuals not present in the logs, reflecting support-mismatch issues in offline RL~\citep{thomas2015hcpi,levine2020offline}. 
Thus execution evidence serves as RLVR-like feedback for artifact governance, with replay-relative rather than future-shift safety claims. 
Appendix~\ref{app:rlvr_threshold_comparison} summarizes the design contrast.

\subsection{Runtime Execution and Maintenance Loop}\label{ssec:runtime}

Runtime handles a query $q$ by executing the latest released workflow or, if none exists, synthesizing a provisional workflow $\hat{\mathcal{W}}$ from task intent and current state. 
It maintains temporary per-run memory $\mathcal{M}$ alongside the authoritative repository $\mathcal{K}$; $\mathcal{M}$ is mutable and non-authoritative, storing transient observations, local repairs, and provisional routines.

Execution is gate-conditioned at the step level. 
For step $s_i$, runtime checks $\gamma_{i,\mathrm{pre}}(x_t)$ on current state $x_t$; if drift causes failure, agents attempt local repair before execution. 
Runtime then follows the fallback cascade $C_i\to P_i\to I_i$ until one realization satisfies both gates. 
If all realizations fail within the retry budget, the run is unresolved and its trace is submitted to maintenance, but $\mathcal{K}$ is unchanged until offline validation accepts a repair or replacement. 
Only accepted repairs update repair evidence. 
Algorithm~\ref{alg:runtime} and Appendix~\ref{app:runtime_system} give the execution procedure, trigger criteria, recovery behavior.

Offline maintenance closes the loop. 
Each execution is distilled into non-authoritative temporary evidence in $\mathcal{M}$ and, when warranted, durable proposals updates. 
This yields a retrieve $\rightarrow$ execute $\rightarrow$ distill $\rightarrow$ promote loop in which runtime proposes changes while maintenance agents, counted in maintenance cost, verify, refactor, de-duplicate, and promote artifacts into new repository versions. 
The thresholds from \S\ref{ssec:adaptive_thresholds} govern workflow release or retirement and step formalization or demotion.

\section{Experiments}\label{sec:experiments}
We evaluate \textsf{SKILL.nb} along three axes: fresh-start task performance on WebArena-Verified and directional transfer on Mind2Web (\S\ref{ssec:main-results}),
repeated-use lifecycle reliability (\S\ref{ssec:lifecycle-dynamics}), and real application-version drift (\S\ref{ssec:gitlab-version-drift}). Supporting these experiments, we show detailed ablation for isolated mechanism contributions (Appendix~\ref{app:component_ablations}) and threshold specialization (Appendix~\ref{app:threshold-ablation}).


\noindent\textbf{Baselines.}
We compare \textsf{SKILL.nb} against baselines targing different capabilitie. We use CodeAct~\citep{wang2024executablecodeactionselicit}, AWM\textsubscript{online}~\citep{wang2024agentworkflowmemory}, and ReasoningBank~\citep{ouyang2025reasoningbank}, representing executable action generation, workflow memory, and retrieved reasoning memory, respectively.

\noindent\textbf{Benchmarks.}
To align with prior workflow- and reasoning-memory evaluations~\citep{wang2024agentworkflowmemory,ouyang2025reasoningbank}, we evaluate on WebArena-Verified~\citep{elhattami2025webarenaverified} and Mind2Web~\citep{dengMind2WebGeneralistAgent2023a}. WebArena-Verified provides 812 tasks across five self-hosted websites. We use its 258-task hard subset for component and threshold ablations. Mind2Web is used for single-round generalization on the cross-task, cross-website, and cross-domain test splits.

\noindent\textbf{Evaluation Protocol.} All methods are re-evaluated in a shared harness for a fair comparison. WebArena-Verified reports 95\% Wilson CIs and uses a two-sided continuity-corrected McNemar test for the main paired success
comparison; Mind2Web reports point estimates because its metrics mix macro-averaged and task-level quantities. Each experiment below provides specific experimental setup and full protocol details are in Appendix~\ref{app:evaluation_protocol}.

\subsection{Benchmark Performance and Generalization}\label{ssec:main-results}

\noindent\textbf{WebArena-Verified.}
We report single-round performance on all 812 WebArena-Verified tasks. Each method starts without task-specific persistent state and builds any repository or memory online during that round. The primary metric is task success rate (SR). Table~\ref{tab:exp_summary} reports overall and per-website SR. We do not compare raw step counts because methods introduce different step types by design.

\begin{table}[ht]
	\centering
	\footnotesize
	\caption{\textsf{SKILL.nb} achieves 53.7\% success on WebArena-Verified, outperforming the next-best baseline by 3.9 points ($p=0.029$). Overall SR is the task-weighted success rate across all 812 tasks (95\% Wilson CI in brackets). Site-specific columns report SR point estimates for each domain. All methods start without preloaded workflow or memory states.}\label{tab:exp_summary}
	\begin{tabular}{lccccccc}
		\toprule
		Method                    & SR (95\% CI) $\uparrow$ & Shopping      & Admin         & Reddit        & GitLab        & Maps          & Multi         \\
		\midrule
		CodeAct                   & 38.3 [35.0, 41.7]       & 39.0          & 44.5          & 55.7          & 31.2          & 33.9          & 10.2          \\
		AWM\textsubscript{online} & 46.4 [43.0, 49.9]       & 48.1          & 49.3          & 68.9          & 40.0          & 45.3          & 5.7           \\
		ReasoningBank             & 49.8 [46.3, 53.2]       & \textbf{52.1} & 55.8          & 68.9          & 45.2          & 42.1          & 11.3          \\
		\textsf{SKILL.nb} (ours)  & \textbf{53.7 [50.3, 57.1]} & 52.0          & \textbf{56.2} & \textbf{70.3} & \textbf{54.4} & \textbf{50.8} & \textbf{17.5} \\
		\bottomrule
	\end{tabular}
\end{table}

\textsf{SKILL.nb} achieves the highest SR at 53.7\%, outperforming ReasoningBank by 3.9 percentage points ($p=0.029$), AWM\textsubscript{online} by 7.3 points, and CodeAct by 15.4 points. The largest site-level gains appear on GitLab (+9.2 points over the next-best baseline) and Maps (+5.5 points). Run-log analysis indicates that GitLab gains stem from more stable access to projects and issues, while Maps gains reflect selective formalization that avoids brittle map and routing interactions. The performance lead on the Multi-site subset (17.5\% vs. 11.3\%) suggests that lifecycle governance is particularly effective for long-horizon tasks. More broadly, these fresh-start gains indicate that even during initial discovery, \textsf{SKILL.nb} benefits from executable formalization that provides more robust state-handling than baselines relying on natural-language reasoning or unstructured memories.

\begin{table}[ht]
	\centering
	\footnotesize
	\caption{\textsf{SKILL.nb} achieves the best overall performance on Mind2Web, leading on all metrics in the cross-website and cross-domain splits. Results provide directional evidence of transfer, especially at the step level. We show element accuracy (EA, $\uparrow$), action F\textsubscript{1} (AF\textsubscript{1}, $\uparrow$), step success rate (SSR, $\uparrow$), and task success rate (SR, $\uparrow$). EA, AF\textsubscript{1} are macro-averaged; SSR, SR are micro-averaged.}\label{tab:mind2web}
	\begin{tabular}{l *{12}{c}}
		\toprule
		                          & \multicolumn{4}{c}{Cross-Task (252)} & \multicolumn{4}{c}{Cross-Website (177)} & \multicolumn{4}{c}{Cross-Domain (912)}                                                                                                                                                          \\
		\cmidrule(lr){2-5} \cmidrule(lr){6-9} \cmidrule(lr){10-13}
		Method                    & EA                                   & AF\textsubscript{1}                     & SSR                                    & SR           & EA            & AF\textsubscript{1} & SSR           & SR           & EA            & AF\textsubscript{1} & SSR           & SR           \\
		\midrule
		CodeAct                   & 46.0                                 & 59.1                                    & 40.3                                   & 3.3          & 39.8          & 45.1                & 31.7          & 1.7          & 35.8          & 37.9                & 31.9          & 1.0          \\
		AWM\textsubscript{online} & 50.0                                 & 56.4                                    & 43.6                                   & 4.0          & 42.1          & 45.1                & 31.6          & 1.6          & 40.9          & 36.3                & 35.5          & 1.5          \\
		ReasoningBank             & \textbf{52.1}                        & 60.4                                    & 44.9                                   & \textbf{4.8} & 44.3          & 52.6                & 34.9          & 3.1          & 40.6          & 41.3                & 36.6          & 1.6          \\
		\textsf{SKILL.nb} (ours)  & 52.0                                 & \textbf{63.2}                           & \textbf{45.0}                          & 4.7          & \textbf{46.8} & \textbf{54.3}       & \textbf{38.1} & \textbf{4.1} & \textbf{44.9} & \textbf{46.0}       & \textbf{39.7} & \textbf{1.9} \\
		\bottomrule
	\end{tabular}
\end{table}

\noindent\textbf{Mind2Web.}
To test zero-shot generalization to unseen environments, we report performance on Mind2Web (Table~\ref{tab:mind2web}). Absolute task success remains low across all methods, but \textsf{SKILL.nb} achieves the best values across all four metrics in the cross-website and cross-domain settings. Given the modest split sizes, we treat the overall pattern across metrics as more informative than any single cell-level gap. On the cross-task split, \textsf{SKILL.nb} closely matches ReasoningBank on task success but leads by 2.8 percentage points on action F$_1$. We interpret these results as directional evidence of transfer at the step level. In the cross-website setting, workflows transfer because high-level intents remain similar despite UI differences; in the cross-domain setting, step-level transfer remains viable because finer-grained interactions---such as search and form filling---recur across the web.


\subsection{Lifecycle Dynamics: Reuse Consistency, Repair, and Regression}\label{ssec:lifecycle-dynamics}

\begin{figure}[H]
	\centering
	\includegraphics[width=\textwidth]{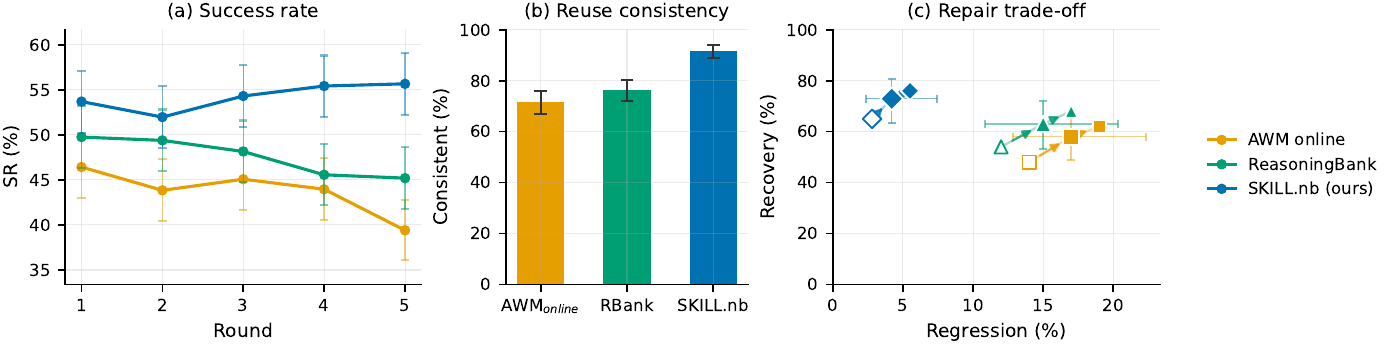}
\caption{\textsf{SKILL.nb} improves task success over repeated rounds, maintains the highest reuse consistency, and optimizes the recovery--regression trade-off. (a)~Task success over five perturbed rounds. (b)~Reuse consistency: fraction of workflows surviving three perturbed re-executions without updates. (c)~Recovery vs. regression under each method’s native update path. Emphasized markers denote repair budget 2 (Table~\ref{tab:repair_ci}). Error bars indicate 95\% Wilson CIs.}\label{fig:lifecycle_sustained}
\end{figure}

Beyond initial success, we investigate whether persistent workflow or memory states benefit agents on recurring tasks. We assess this through two lifecycle tests. The first measures round-level reuse (Figure~\ref{fig:lifecycle_sustained}(a)): methods execute five WebArena-Verified rounds, carrying persistent states across iterations while per-task environment states and transient contexts reset. Each round applies a uniform perturbation protocol across all methods: tasks are reshuffled, starting URLs are varied, and intent templates are paraphrased while preserving core slots and intents. The second test isolates artifact reuse under perturbation before and after update. We first snapshot initially successful states and re-execute them three times without updates to measure reuse consistency (Figure~\ref{fig:lifecycle_sustained}(b)). Failed snapshots then enter each method's native update path, where we measure recovery and update-induced regression (Figure~\ref{fig:lifecycle_sustained}(c)). Appendix~\ref{app:evaluation_protocol} gives the full protocol.

\textsf{SKILL.nb} is the only method to improve over repeated rounds, rising from 53.7\% to 55.7\% by round~5, while baselines decline by 4.6--7.0 percentage points (Figure~\ref{fig:lifecycle_sustained}(a)). This advantage reflects superior artifact-level stability: \textsf{SKILL.nb} maintains 91.7\% reuse consistency, above ReasoningBank (76.2\%) or AWM\textsubscript{online} (71.6\%; Figure~\ref{fig:lifecycle_sustained}(b)). When artifacts fail and enter native update paths, \textsf{SKILL.nb} optimizes the recovery--regression trade-off (72.9\%/4.2\%), outperforming ReasoningBank (63.0\%/15.0\%) and AWM\textsubscript{online} (58.0\%/17.0\%) at budget~2 (Figure~\ref{fig:lifecycle_sustained}(c)). 
These results indicate that while ungated memories often absorb regressive updates, \textsf{SKILL.nb}’s validation-gated promotion ensures that persistent workflow state remains stable under repeated perturbation and repair.

\begin{wraptable}{r}{0.50\textwidth}
	\vspace{-1.0em}
	\footnotesize
	\setlength{\tabcolsep}{4pt}
	\caption{Main mechanism ablation summary on the 258-task WebArena-Verified hard subset. SR and regression are percentages; token cost is \textsf{SKILL.nb}-internal maintenance/update tokens per success. Full diagnostics are in Appendix~\ref{app:component_ablations}.}
	\label{tab:main_mechanism_ablation}
	\begin{tabular}{lccc}
		\toprule
		Variant & SR (\%) $\uparrow$ & Regression (\%) $\downarrow$ & Tok./succ. $\downarrow$ \\
		\midrule
		\textsf{SKILL.nb} & \textbf{38.4} & \textbf{3.3} & \textbf{36k} \\
		NL-only & 33.3 & 8.8 & 57k \\
		Code-only & 31.0 & 14.7 & 42k \\
		No gates & 32.6 & 18.6 & 45k \\
		No demote & 34.1 & 12.4 & 47k \\
		\bottomrule
	\end{tabular}
	\vspace{-1.0em}
\end{wraptable}

\noindent\textbf{Mechanism and overhead diagnostics.}
Table~\ref{tab:main_mechanism_ablation} shows component ablations on the WebArena-Verified hard subset. The full system has the highest point SR and lowest \textsf{SKILL.nb}-internal maintenance/update token cost per success. Removing executable formalization or disabling fallback lowers success, while removing gates increases update-induced regression. In a separate three-round threshold-policy ablation (Figure~\ref{fig:threshold_ablation}), group-specialized thresholds give the best success--regression--maintenance trade-off: by round~3, \textsf{SKILL.nb} reaches 38.3\% success with 3.3\% regression, whereas loose fixed thresholds over-promote and fall to 27.1\% success with 22.0\% regression. 
Maintenance overhead amortizes over reuse, with token usage per successful task falling to 69.2\% of its round-1 value by round~5. 
Full ablations, threshold curves, and cost curves are in Appendices~\ref{app:component_ablations}, \ref{app:threshold-ablation}, and~\ref{app:maintenance-cost}.

\subsection{Adapting to Real-World Environment Drift}\label{ssec:gitlab-version-drift}
We next test whether persistent state remains useful under real interface drift.
We migrate all 180 GitLab tasks from WebArena-Verified from GitLab~15 to two target versions: GitLab~16, the first major UI-changing release relative to the original deployment, and GitLab~18.9, the latest available version. Appendix Figure~\ref{fig:gitlab_ui_drift_example} and Table~\ref{tab:gitlab_history_dom_drift} illustrate the corresponding UI and DOM drift. Unlike synthetic perturbations, these updates change DOM structure, selectors, and page flow while preserving user intent.
Each method is evaluated in five conditions: fresh-start runs on GitLab~15,
16, and 18, plus frozen-state reuse on GitLab~16 and 18 using the
repository or memory built on GitLab~15. In the frozen-state conditions, we
restore the GitLab~15 repository or memory snapshot before each target-version
task and discard durable updates afterward. Thus success measures whether
persistent state learned on the source deployment helps or harms execution after
migration, rather than cumulative relearning on the target version. Full protocol
details are in Appendix~\ref{app:gitlab-version-drift}.

\begin{wrapfigure}{r}{0.48\textwidth}
    \centering
    \includegraphics[width=\linewidth]{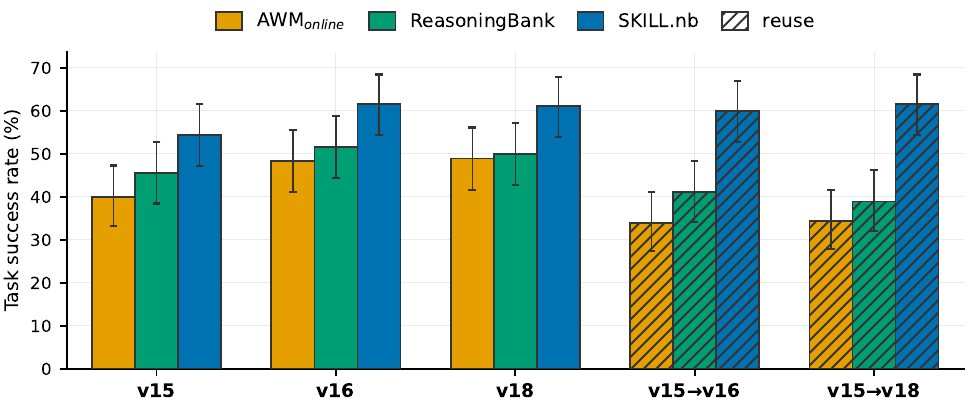}
	\caption{\textsf{SKILL.nb} avoids negative transfer from stale procedural state
under real GitLab version drift. The x-axis shows fresh-start runs on
GitLab~15, 16, and 18, followed by frozen-state reuse from
15$\rightarrow$16 and 15$\rightarrow$18 Within each condition, orange,
green, and blue bars denote AWM\textsubscript{online}, ReasoningBank, and
\textsf{SKILL.nb}. Hatched bars denote frozen repository or memory reuse. Error
bars indicate 95\% Wilson CIs.}
	\label{fig:gitlab_drift}
\end{wrapfigure}

Newer GitLab versions are not intrinsically harder under fresh-start
execution: all methods improve on newer versions.
Qualitative log inspection suggests that newer GitLab versions expose more stable
DOM locators for the Playwright-based harness. They may also better match recent
LLM web-navigation related pre-training knowledge. 

Persistent-memory baselines suffer negative transfer from frozen source-version
state. On GitLab~18, AWM\textsubscript{online} drops from 88/180 fresh-start
successes to 62/180 frozen-state successes (48.9\% to 34.4\%), a 14.4-point
degradation. ReasoningBank similarly drops from 90/180 to 70/180 successes
(50.0\% to 38.9\%), an 11.1-point degradation. Qualitative inspection suggests
that frozen memories often retain interface-specific assumptions, such as brittle
click identifiers or page-geometry-dependent procedures, that no longer match the
migrated DOM.
In contrast, \textsf{SKILL.nb} preserves target-version performance under
frozen-state reuse. On GitLab~18, it obtains 110/180 fresh-start successes and
111/180 frozen-state successes (61.1\% vs.\ 61.7\%). The same pattern holds on
GitLab~16, where frozen-state reuse reaches 108/180 successes compared with
111/180 fresh-start successes. This result is consistent with gate-conditioned
selective formalization: when an executable realization no longer satisfies its
pre/postcondition gates, runtime can fall back from code to the preserved NL
procedure or step intent. Thus, under real GitLab version drift,
\textsf{SKILL.nb} avoids the old-state degradation observed in persistent-memory
baselines.

\section{Conclusion and Limitations}\label{sec:conclusion}

We presented \textsf{SKILL.nb}, a lifecycle-governance framework for durable web-agent workflow artifacts. 
Its core mechanism, \emph{selective formalization}, uses execution-grounded evidence to decide when workflows are released or retired, and when steps remain NL-guided, become executable, or are demoted after repair. 
Versioned notebooks provide the governed artifact: they bind NL procedures, executable realizations, validation gates, and maintenance history into a reusable object. 
Across WebArena-Verified, Mind2Web, and GitLab version drift, \textsf{SKILL.nb} improves single-round task performance, repeated-use reliability, repair recovery, and robustness to stale persistent state. 
These results suggest that durable agent artifacts should be treated not only as memories, but as lifecycle-managed objects whose promotion, repair, and retirement are governed by execution evidence.

\noindent\textbf{Limitations.}
\textsf{SKILL.nb}'s lifecycle decisions are bounded by logged execution evidence. 
The Wilson-UCB feasibility check bounds estimated violation rates on threshold-estimation cases, not arbitrary future workloads or interface shifts. 
The method also depends on reliable gates, metadata quality, group assignment, and recurring task structure; gate errors or sparse recurrence can lead to invalid executions or conservative pooled behavior. 
Our maintenance-cost proxy counts LLM inference tokens rather than total operational cost, such as wall-clock latency, storage, or human review. 
Finally, our evaluation studies reuse and lifecycle correction in controlled benchmark environments, including single-application GitLab version migration, which may differ from long-term production deployment. 
Appendix~\ref{app:limitations_scope} discusses these limitations in detail.

\bibliographystyle{plainnat}
\bibliography{references}

\clearpage
\appendix
\section{Method Details}\label{app:method_details}

\subsection{From Provisional Trace to Released Workflow Artifact}\label{app:artifact_example}

\begin{figure}[H]
	\centering
	\includegraphics[width=0.98\linewidth]{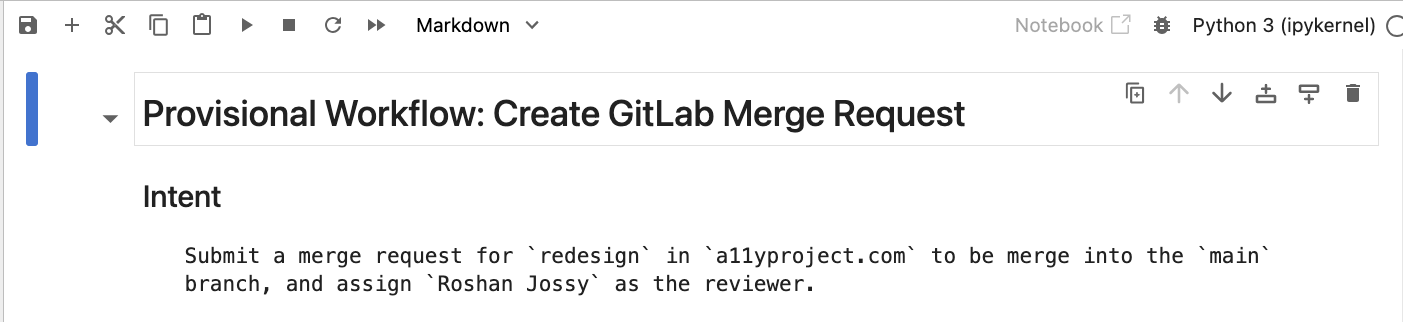}
	\vspace{0.25em}
	{\small\textbf{(a)} Cell 1: provisional header cell with task-specific values.}
	\vspace{0.8em}

	\includegraphics[width=0.98\linewidth]{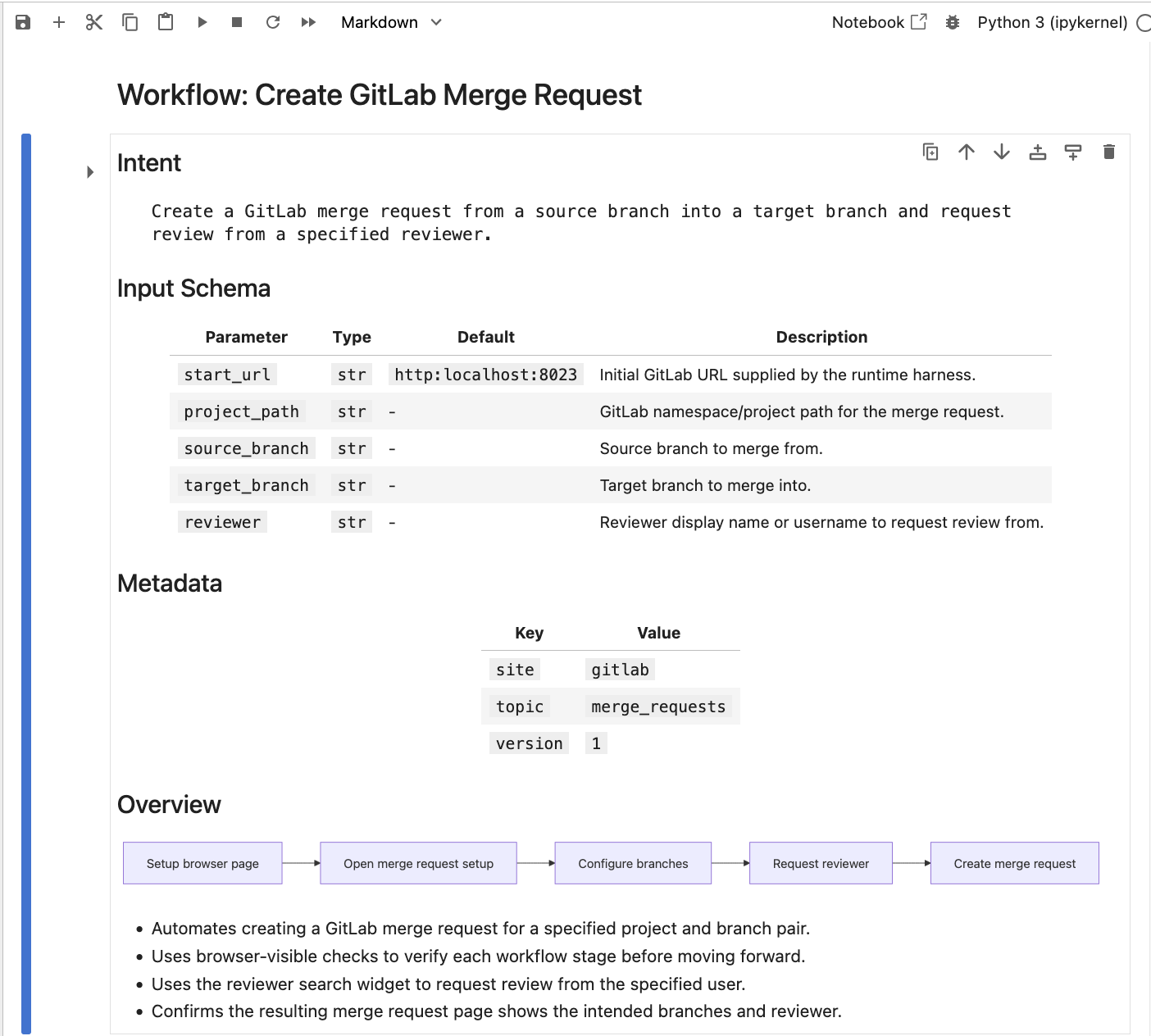}
	\vspace{0.25em}
	{\small\textbf{(b)} Cell 1: released header cell with reusable inputs and metadata.}
	\caption{Header-cell generalization for the GitLab merge-request lifecycle example. The provisional notebook records the concrete task request in Cell~1, while the released artifact keeps the reusable workflow description, input schema, and metadata so future requests can instantiate the same procedure with different branches, projects, and reviewers.}
	\label{fig:header_cell_generalization_example}
\end{figure}

This appendix illustrates how a task-specific provisional notebook is converted into a reusable released \textsf{SKILL.nb} workflow artifact. The example uses a GitLab merge-request task to show how concrete task values are lifted into an input schema, how notebook cells are aligned to workflow steps, and how execution evidence supports validation and promotion. Runtime gates use browser-observable state and task-provided expected values; notebook outputs, logs, screenshots, and cell status are used by offline maintenance during artifact review. The example does not use hidden benchmark labels or evaluator outputs.

The concrete example is Task~668, a GitLab task starting from \url{http://localhost:8023}. The user request is to submit a merge request for source branch \texttt{redesign} in project \texttt{a11yproject.com}, merge it into the \texttt{main} branch, and assign \texttt{Roshan Jossy} as the reviewer. This task instance supplies concrete values for the source branch, target branch, project, and reviewer. During maintenance, these values are lifted into an input schema, while the task-specific request is rewritten as a generic merge-request workflow intent. This keeps the artifact reusable without storing the benchmark task ID as part of the executable workflow.

Figure~\ref{fig:header_cell_generalization_example} shows the first promotion transformation. The task-specific request supplies concrete inputs: source branch \texttt{redesign}, target branch \texttt{main}, source project \texttt{a11yproject.com}, and reviewer \texttt{Roshan Jossy}. Maintenance distills this instance into a reusable header cell that records the workflow description, input schema $X$, and workflow metadata $M^W$ for retrieval.

\begin{figure}[H]
	\centering
	\includegraphics[width=0.98\linewidth,height=0.40\textheight,keepaspectratio]{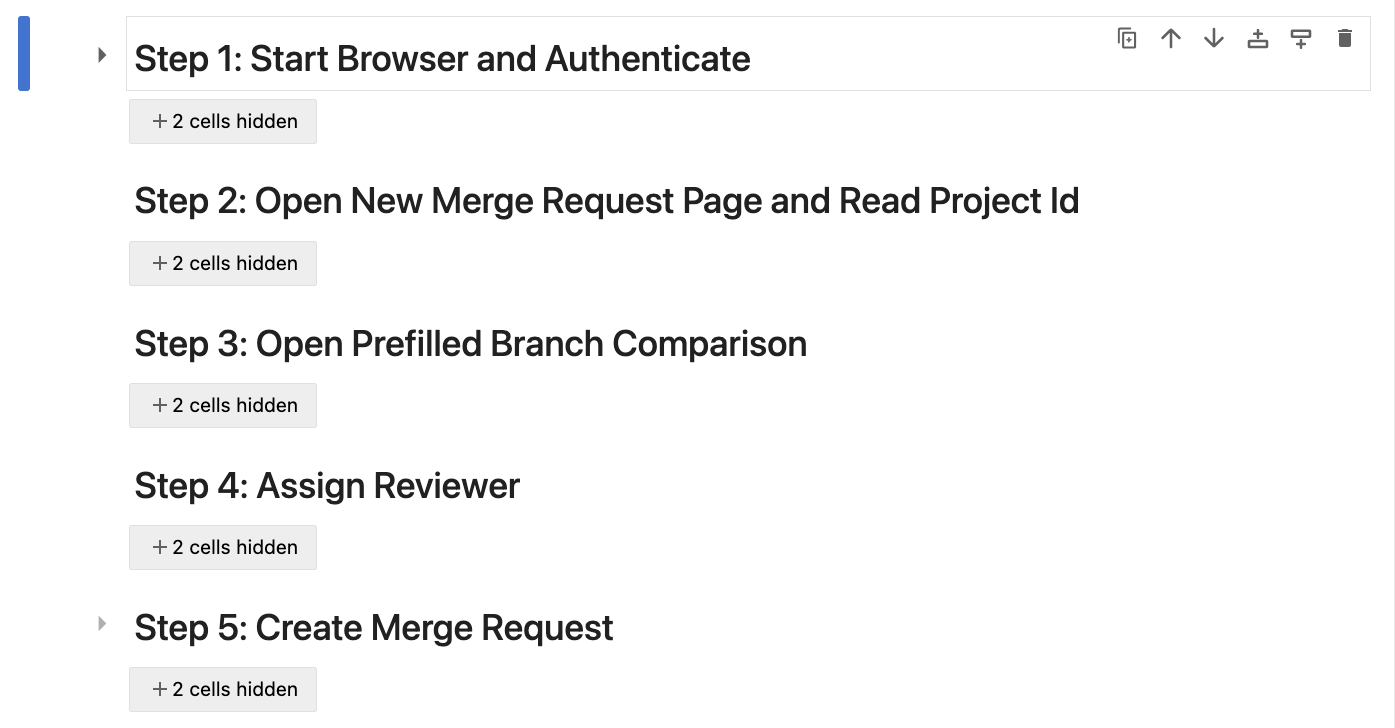}
	\vspace{0.25em}
	{\small\textbf{(a)} Cells 2--N: provisional workflow steps.}
	\vspace{0.8em}

	\includegraphics[width=0.98\linewidth,height=0.40\textheight,keepaspectratio]{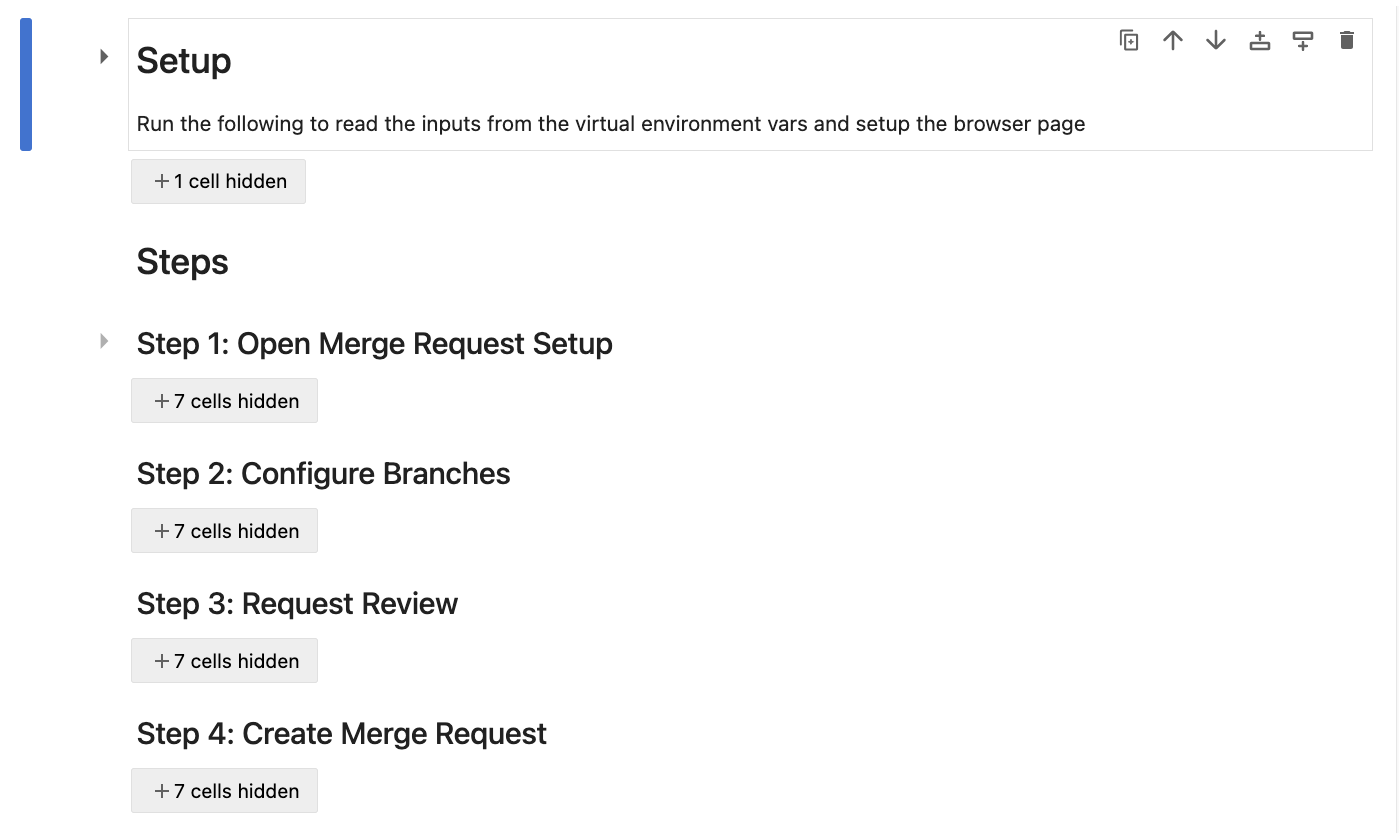}
	\vspace{0.25em}
	{\small\textbf{(b)} Cells 2--N: released workflow steps with executable gates.}
	\caption{Step-cell transformation for the GitLab merge-request lifecycle example. Cells~2--N contain the executable workflow steps: setup, browser actions, checks, and submission. The released artifact preserves the reusable step structure while replacing task-specific traces with parameterized inputs and validation gates.}
	\label{fig:step_cell_transformation_example}
\end{figure}

\begin{figure}[H]
	\centering
	\includegraphics[width=0.98\linewidth,height=0.43\textheight,keepaspectratio]{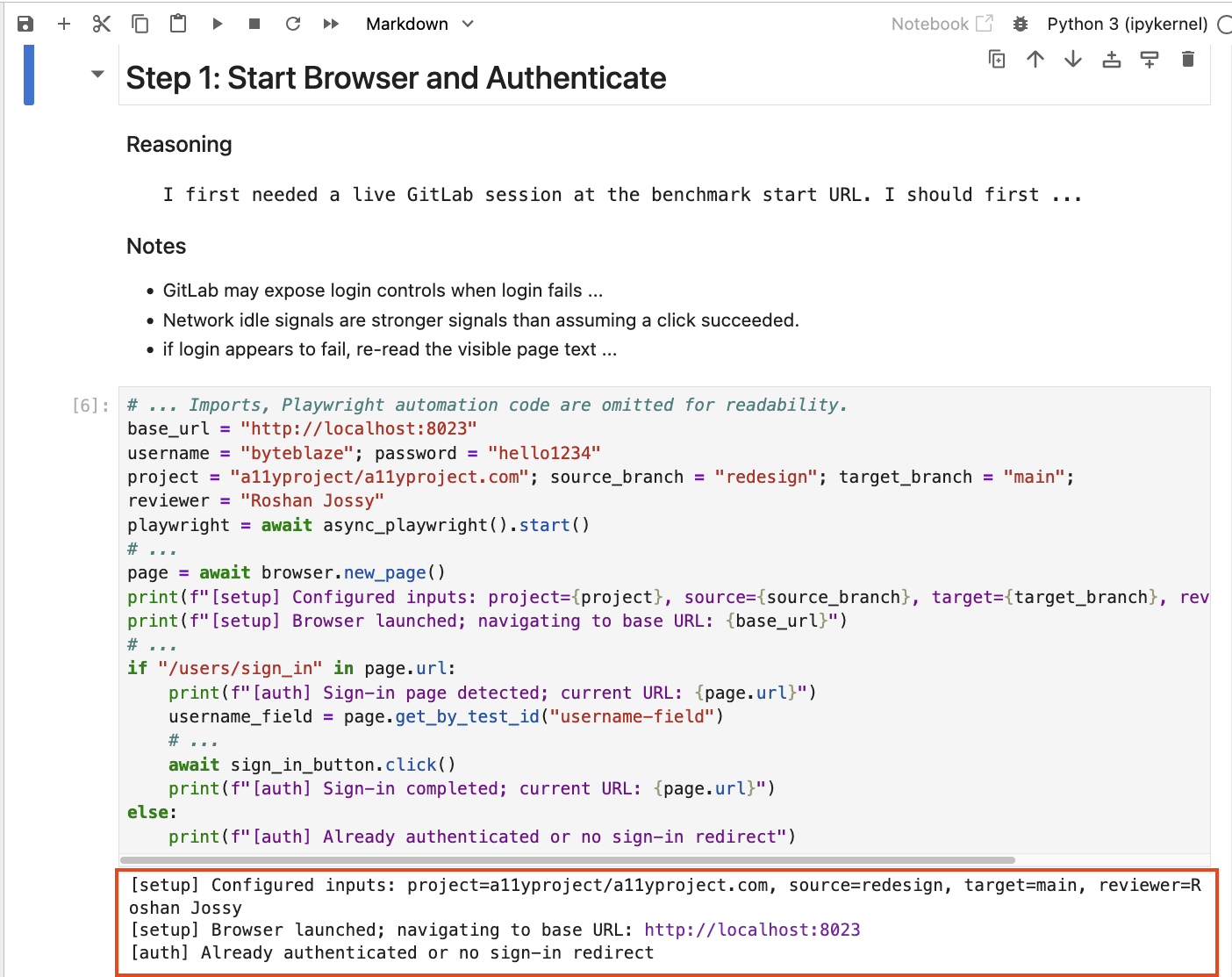}\par
	\vspace{0.25em}
	{\small\textbf{(a)} Cells 2--3: provisional Step 1 with setup, checks, and red-boxed execution log.}
	\vspace{0.8em}

	\includegraphics[width=0.98\linewidth,height=0.43\textheight,keepaspectratio]{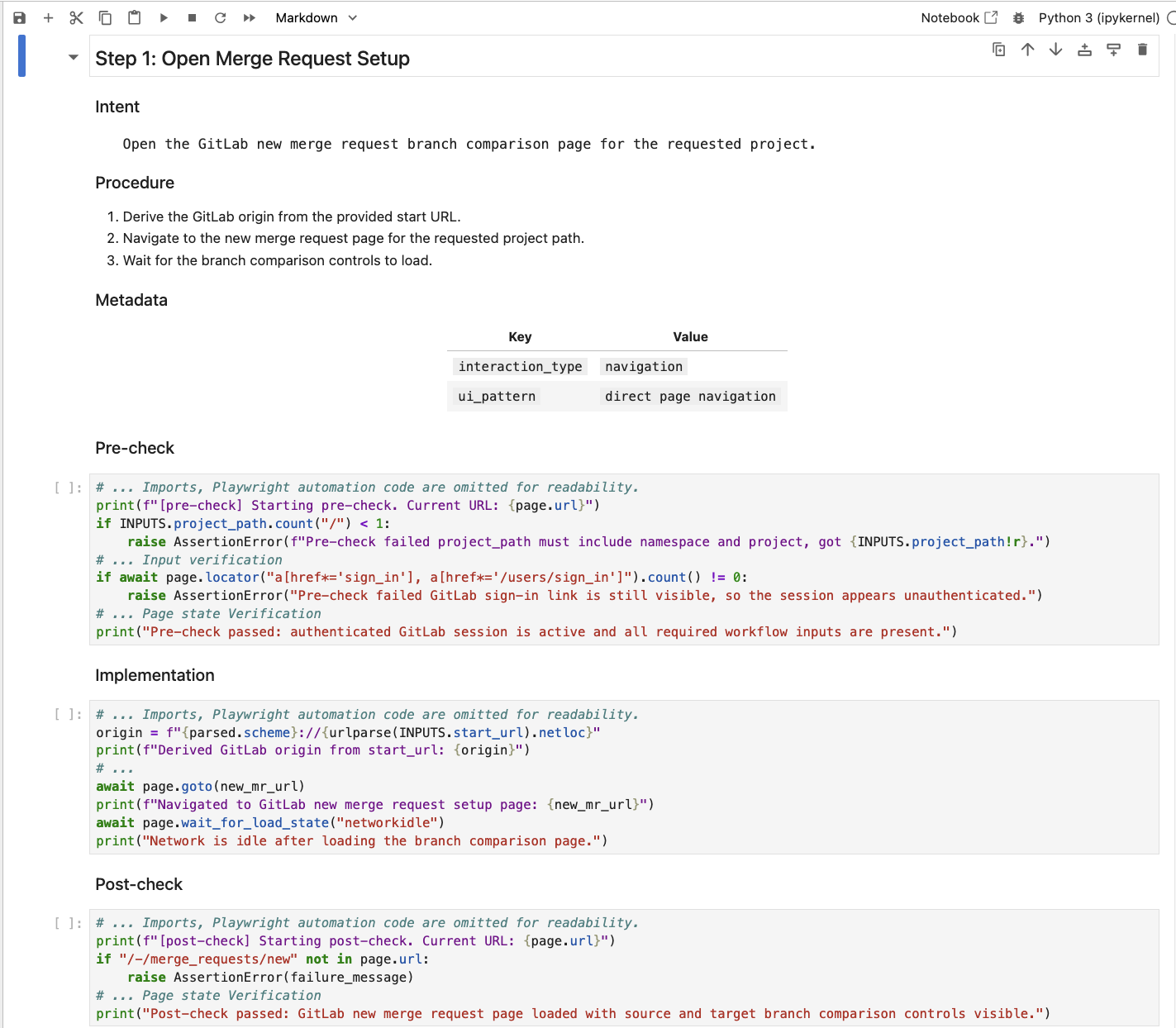}\par
	\vspace{0.25em}
	{\small\textbf{(b)} Cells 2--8: released Step 1 with parameterized setup and gates.}
	\caption{Expanded Step~1 transformation for the GitLab merge-request lifecycle example. The provisional step in (a), from the provisional notebook, is formalized into (b), from the released notebook, as $s_1=\langle I_1,P_1,C_1,\Gamma_1,M_1^S\rangle$ with intent, procedure, executable setup, gates, and metadata.}
	\label{fig:step_1_transformation_example}
\end{figure}

After the provisional workflow executes, maintenance uses notebook evidence to align traces to the step cells in Figure~\ref{fig:step_cell_transformation_example}, then expands individual stages as in Figure~\ref{fig:step_1_transformation_example} to decide which steps should be formalized as executable cells with gates. In this example, trace evidence supports promotion from $y(\mathcal{W}_v)=\texttt{provisional}$ to $y(\mathcal{W}_v)=\texttt{released}$, while validated executable cells change the corresponding step indicators from $z_i=0$ to $z_i=1$. Each formalized step stores a local intent $I_i$, natural-language procedure $P_i$, executable realization $C_i$, executable pre/post gates $\Gamma_i$, and metadata $M_i^S$. The gates can check the current GitLab context, branch fields, reviewer selection, navigation state, and successful form submission. If later repair evidence shows that the workflow has become too costly or unstable, maintenance can move the same workflow lineage to $y(\mathcal{W}_v)=\texttt{retired}$ while retaining it for rollback and analysis. Because creating a merge request is non-idempotent, maintenance validation controls promotion rather than blindly rerunning the same submission.

\subsection{Notebook Evidence and Debugging Affordances}\label{app:notebook_affordances}

The previous subsection used Task~668 to illustrate artifact promotion. The next figures show additional notebook affordances used by maintenance across workflows: cell-attached screenshots, interactive debugging state, localized failure evidence, and heterogeneous execution cells.

\begin{figure}[H]
	\centering
	\includegraphics[width=0.98\linewidth,height=0.42\textheight,keepaspectratio]{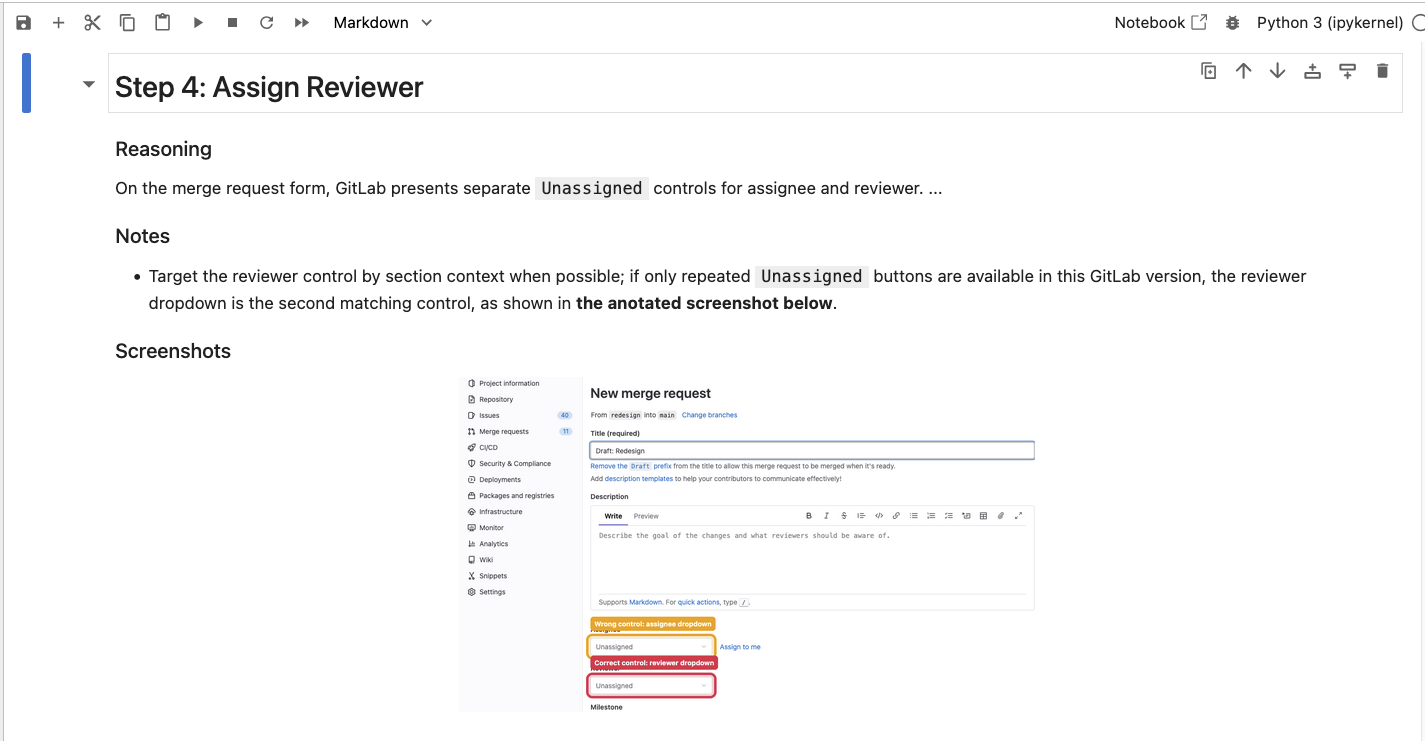}
	\caption{Cell-attached screenshot evidence for dynamic form handling. When provisional creation or execution encounters ambiguous controls, such as repeated \texttt{Unassigned} dropdowns for assignee and reviewer, the agent attaches the observed UI to the relevant cell. The embedded screenshot is cropped for readability; the actual notebook can link to the full-page screenshot used by maintenance to formalize reusable gates.}
	\label{fig:cell_attached_screenshot_evidence}
\end{figure}

\begin{figure}[H]
	\centering
	\includegraphics[width=0.98\linewidth,height=0.42\textheight,keepaspectratio]{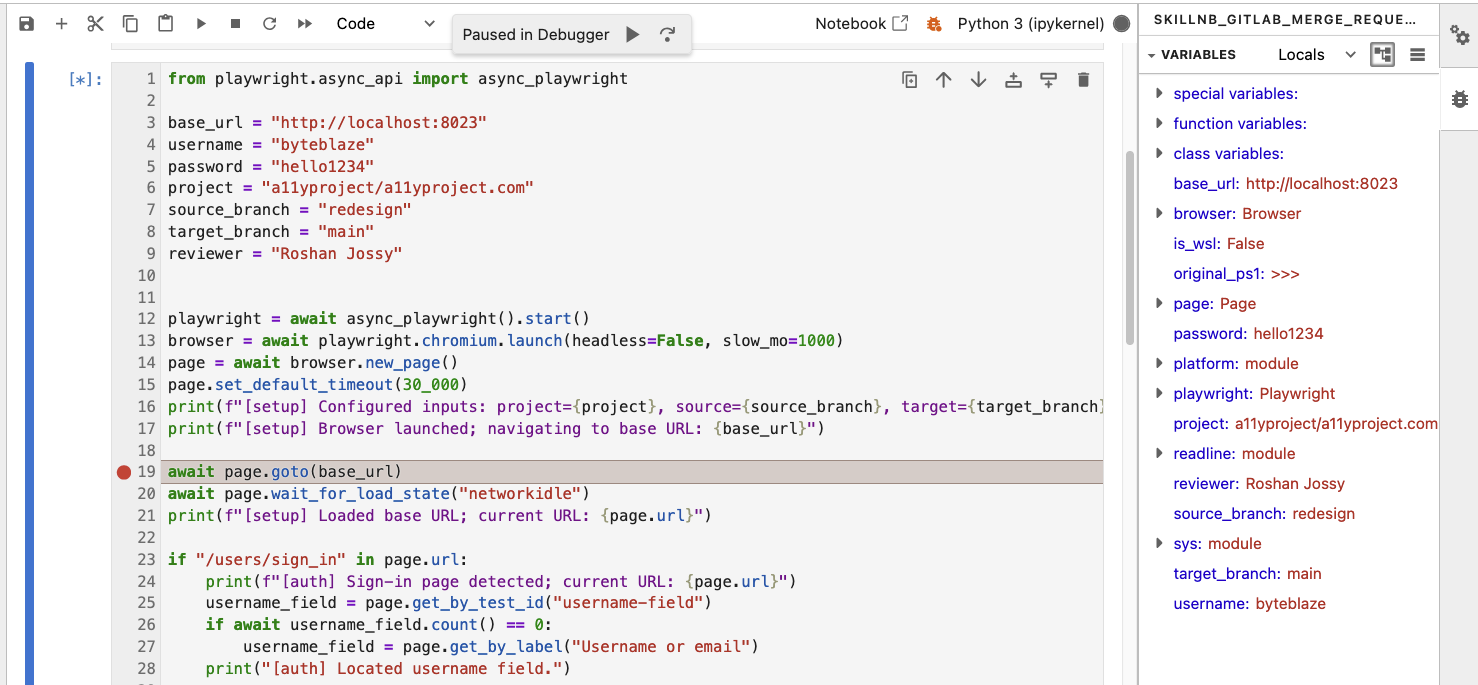}
	\caption{Interactive debugging support from the notebook representation. Because workflow realizations live in executable Jupyter cells, an agent or maintenance process can pause at a breakpoint, step through browser automation code, inspect task inputs and Playwright page state, and turn observed failures into more reliable executable gates.}
	\label{fig:jupyter_debug_step_evidence}
\end{figure}

\begin{figure}[H]
	\centering
	\includegraphics[width=0.98\linewidth,height=0.42\textheight,keepaspectratio]{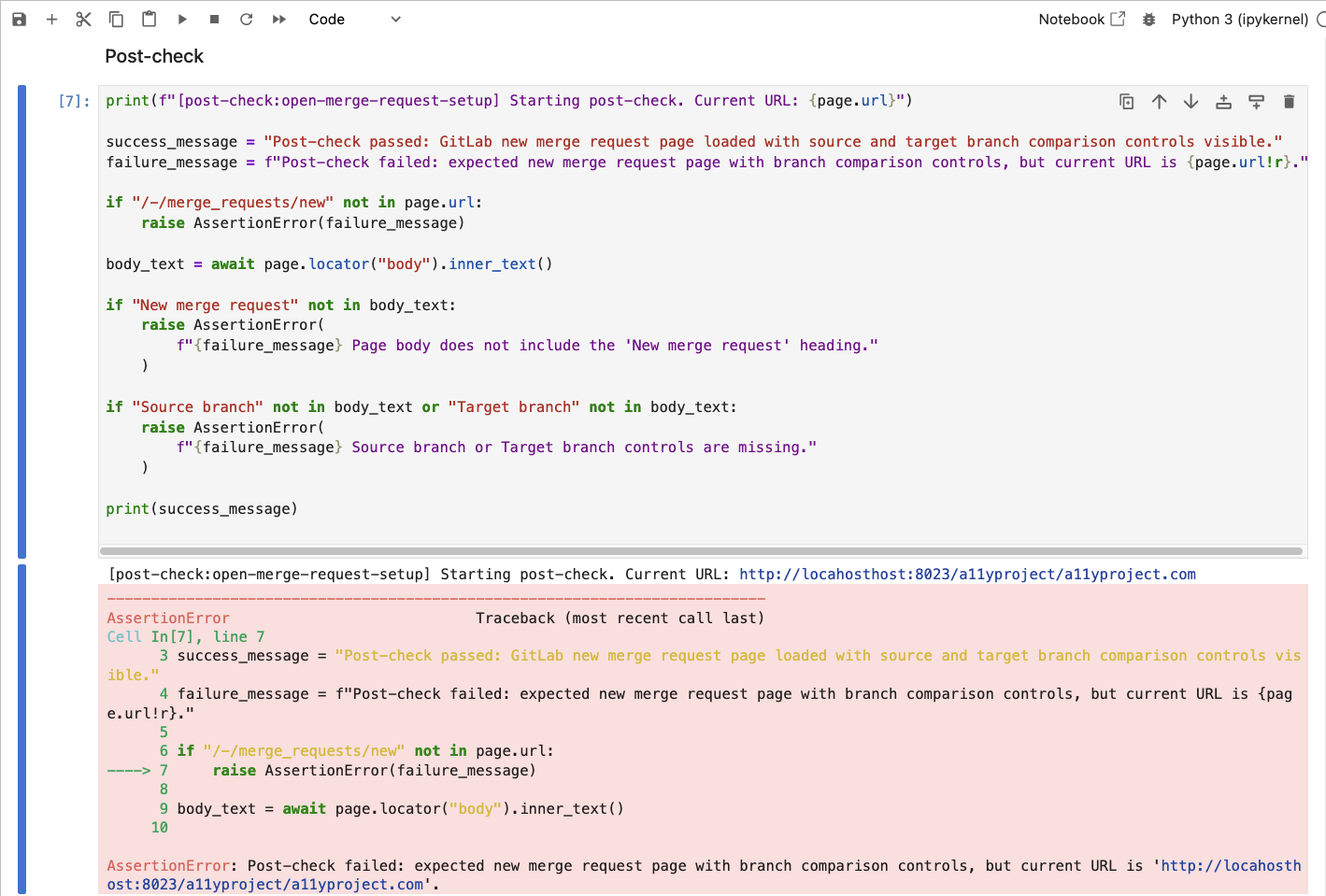}
	\caption{Cell-local failure evidence in a provisional notebook. When a gate fails, Jupyter stores the printed context, exception, and traceback directly under the failing cell, so an agent can retrieve the relevant code, state, and error in one localized artifact. A multi-component logging system could provide similar information, but would add logging infrastructure and synchronization points outside the workflow artifact.}
	\label{fig:cell_local_failure_log}
\end{figure}

\begin{figure}[H]
	\centering
	\includegraphics[width=0.98\linewidth,height=0.42\textheight,keepaspectratio]{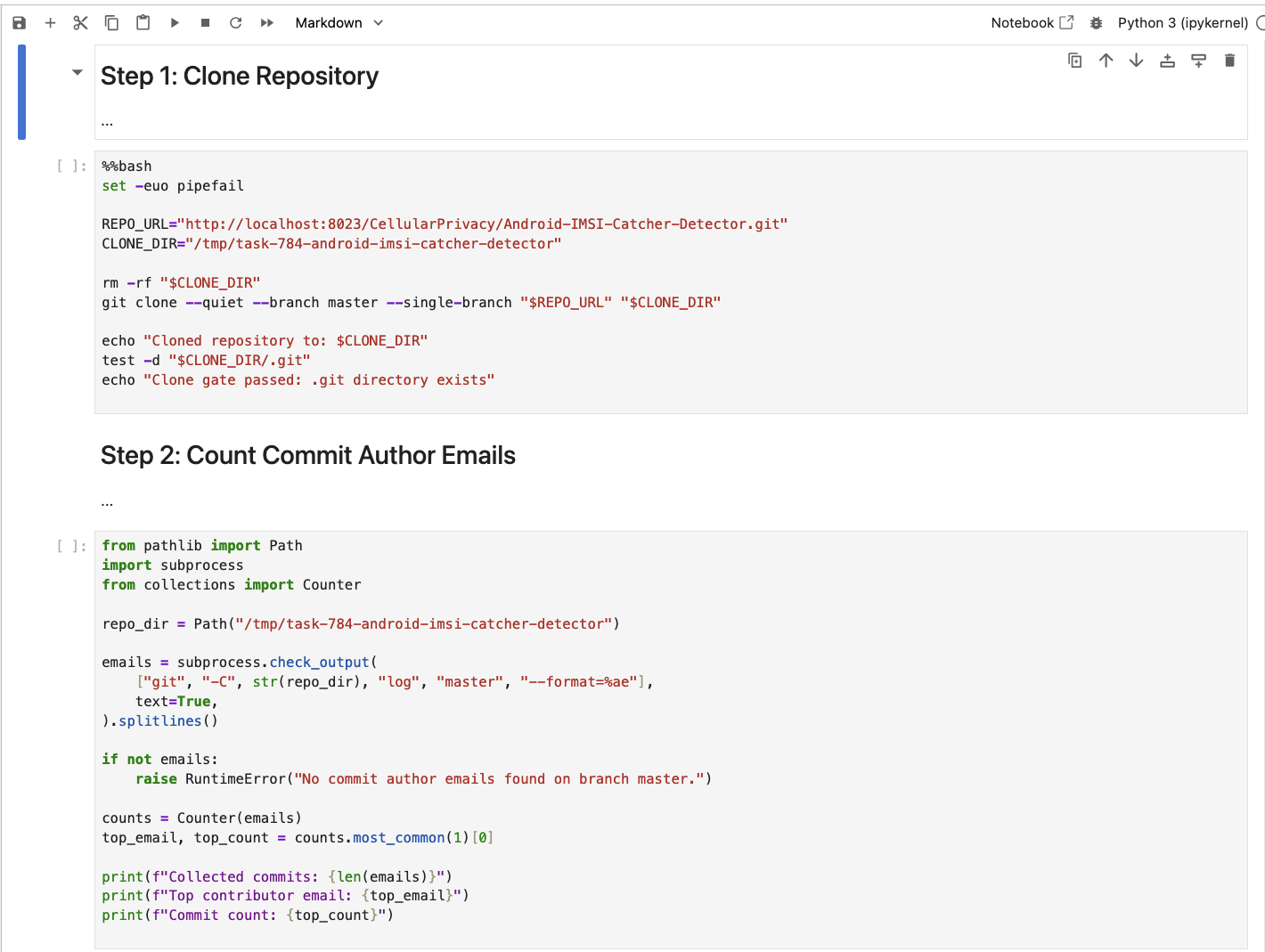}
	\caption{Mixed-language workflow cells in a provisional notebook. For Task~784, the agent can use a Bash cell to clone the GitLab repository \texttt{CellularPrivacy/Android-IMSI-Catcher-Detector}, then use Python cells to analyze commits on branch \texttt{master} and identify the contributor email. This lets each step use the most suitable tool while preserving the full workflow trace in one artifact.}
	\label{fig:mixed_language_notebook_cells}
\end{figure}

Although the main walkthrough uses Task~668, the notebook representation also supports heterogeneous execution cells. Figure~\ref{fig:mixed_language_notebook_cells} shows a separate Task~784 example using Bash and Python cells in the same artifact.

\subsection{\textsf{SKILL.nb} as an Extension of \texttt{SKILL.md}}\label{app:skill_md_vs_nb}

Agent Skills define a lightweight convention for packaging agent capabilities as folders with a \texttt{SKILL.md} file, optional scripts, references, and assets.\footnote{\url{https://agentskills.io/home}} A \textsf{SKILL.nb} artifact can be viewed as a natural extension of this convention rather than a separate abstraction. It preserves the same core ingredients: a human-readable description, procedural instructions, optional executable resources, and supporting artifacts. The key difference is the execution contract. A \texttt{SKILL.md} file may include code blocks, but the markdown artifact alone does not guarantee that an agent executes those blocks exactly as written. Unless an external client or harness enforces verbatim execution, the agent may adapt the snippet, use it as guidance, or ignore it.

\textsf{SKILL.nb} makes this boundary explicit by placing instructions, executable cells, gates, observed outputs, and local evidence in one versioned notebook object. This does not remove the need for task-level execution controls, but it gives maintenance a single auditable artifact for checking what code was intended to run, where validation occurred, and what evidence was produced. The choice also avoids introducing a bespoke trace format: notebooks are a standard and widely supported representation for interleaving markdown, code, metadata, and outputs, and are therefore a familiar substrate for existing tools and LLM-backed agents. In our implementation, the loader supports progressive disclosure: it first reads the leading markdown cell and loads later cells only when execution, validation, or evidence inspection is needed.

Our implementation uses this continuity directly. It overloads the existing skill-loading path so that a notebook artifact can be exposed through the same discovery interface as a \texttt{SKILL.md} skill. At discovery time, the loader extracts the notebook-level name, intent, and description from the leading markdown cell. At activation time, it reads the relevant notebook cells and attached resources. This makes the notebook representation compatible with the same discovery pathway in our implementation while retaining executable cell boundaries, validation gates, and cell outputs.

\begin{figure}[H]
\centering
\includegraphics[width=0.98\linewidth,height=0.75\textheight,keepaspectratio]{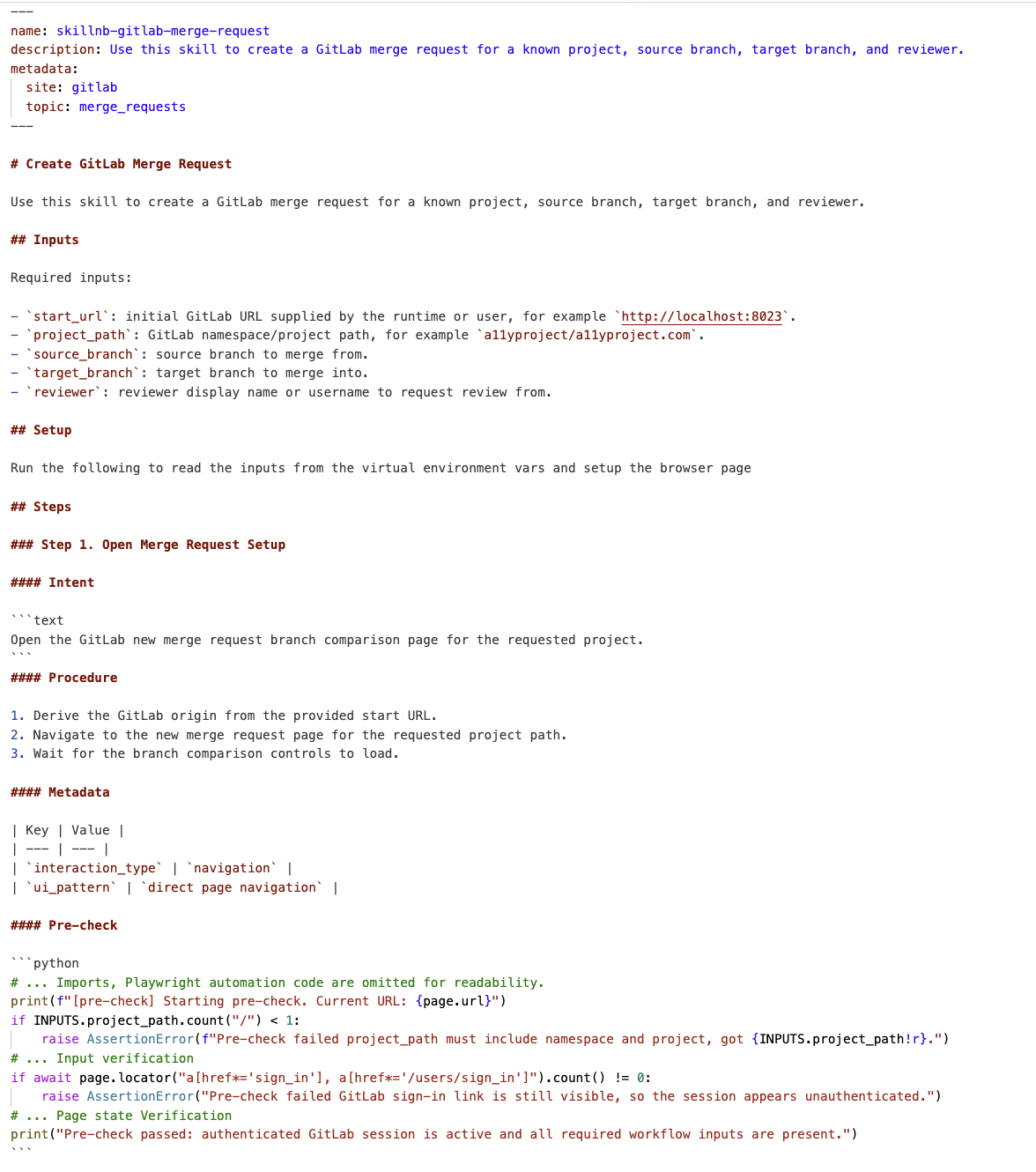}
\caption{Markdown skill representation of Figure~\ref{fig:step_1_transformation_example}(b), Cells~2--8: released Step~1 with parameterized setup and gates. The \texttt{SKILL.md} version stores the same GitLab merge-request workflow as front matter, inputs, natural-language intent and procedure, metadata, and code snippets in a markdown document, whereas the corresponding \textsf{SKILL.nb} artifact stores these units as executable notebook cells with validation gates and cell-local outputs.}
\label{fig:skill_md_vs_nb}
\end{figure}

Figure~\ref{fig:skill_md_vs_nb} shows the corresponding \texttt{SKILL.md} representation for the released Step~1 notebook artifact in Figure~\ref{fig:step_1_transformation_example}(b), while Table~\ref{tab:skill_md_vs_nb} summarizes the design differences.

The comparison is representational rather than a claim that notebooks are intrinsically safer. Execution still depends on the surrounding client and task environment. The benefit of \textsf{SKILL.nb} is that the executable cells, gates, outputs, screenshots, and logs are stored in one artifact, so offline maintenance can audit the intended realization and the evidence used to validate it.

 \begin{table}[H]
\centering
\small
\caption{Comparison between conventional \texttt{SKILL.md} artifacts and \textsf{SKILL.nb} notebook artifacts.}
\label{tab:skill_md_vs_nb}
\begin{tabularx}{\linewidth}{@{}l X X@{}}
\toprule
\textbf{Dimension} & \textbf{\texttt{SKILL.md}} & \textbf{\textsf{SKILL.nb}} \\
\midrule
Discovery & Name and description are read from markdown front matter or leading prose. & Name, intent, and description are read from the leading notebook markdown cell. \\
Progressive disclosure & Full instructions are loaded when the skill is activated. & Later cells are loaded only when the workflow, code, gates, or logs are needed. \\
Procedural guidance & Natural-language instructions are primary. & Natural-language cells remain primary and are interleaved with executable cells. \\
Execution contract & Code blocks are instructions unless an external client enforces verbatim execution. & Executable cells, gates, and outputs are explicit artifact units available for audit and maintenance. \\
Execution evidence & External logs or traces must be attached separately. & Cell outputs, screenshots, and validation logs can live in the artifact. \\
Debugging & Inspecting failures may require separate traces, terminals, or logging systems. & The agent can inspect cell outputs, screenshots, and validation logs locally, without depending on an external logging system. \\
Implementation path & Native format for many skills-compatible clients. & Implemented as an overload of the same skill-loading path, with notebook parsing at activation. \\
Overhead & Compact plain markdown. & JSON plus cell metadata and outputs, trading size for richer execution structure. \\
\bottomrule
\end{tabularx}
\end{table}

The notebook representation has overhead. A notebook is stored as JSON rather than plain markdown, and it carries cell metadata and outputs. This increases artifact size and may make raw diffs less compact than a short \texttt{SKILL.md} file. We view this as an engineering tradeoff. The JSON format is standardized, widely supported, and easy for modern agents to parse. In return, the artifact can combine natural-language procedure, executable code in multiple languages, screenshots, validation logs, and rerun boundaries in one versioned object. This also improves local debugging: an agent can inspect the recorded outputs and validation logs in the same artifact, without requiring a separate tracing or observability system.

\subsection{Artifact Safety and Reproducibility Scope}\label{app:artifact_safety}

The workflow artifact schema records intents, procedures, executable cells, gates, metadata, and version links. It does not include credentials, hidden evaluator labels, or benchmark final answers as durable workflow fields. Runtime traces and local repair memories are non-authoritative until offline maintenance promotes a validated workflow version into $\mathcal{K}$.

In our evaluation, executable notebook cells are run under the surrounding task or benchmark execution controls. We therefore treat sandboxing, dependency pinning, credential scoping, and trace redaction as deployment and evaluation-harness controls rather than as guarantees provided by the lifecycle policy itself. Trace redaction, including redaction of stored visual state when it contains user-provided secrets or credentials, is part of the surrounding deployment harness and outside the algorithmic claims of the lifecycle policy. The method claims here rely on versioned artifacts, deterministic gate checks, offline validation, and rollback-addressable repository state.

\subsection{Validation and Evidence Signals}\label{app:validation_signals}

Table~\ref{tab:validation_signals} summarizes the signals used by \textsf{SKILL.nb}. Runtime gates are deterministic predicates over browser-observable state instantiated with task-provided expected values. They can inspect DOM structure, visibility, URLs, form values, page text, and counts of relevant page objects. Gates do not call benchmark evaluators, hidden success labels, or final-answer oracles.

Offline validation is broader than runtime gating. It may use logged traces, cached regression checks, maintenance review, or calibration-split benchmark labels when such labels are available. These labels are offline-only: they are never available to the runtime controller, and final evaluation tasks are disjoint from the calibration logs and labels used for threshold analysis. Final benchmark trajectories, hidden labels, and task success outcomes are not used to estimate thresholds, assign groups, accept repairs, or promote workflows before evaluation.

A repair is accepted only when it passes the triggering trace's deterministic pre/post gates and offline maintenance validation for the affected workflow version. Accepted repairs update the workflow-version repair count used for demotion and the workflow-lineage repair burden used for retirement. A threshold violation is an offline calibration outcome where a candidate threshold would admit a lifecycle action whose downstream validation loss exceeds $J_{\mathrm{perf}}^{\mathrm{ref}} + \epsilon$.

\begin{table}[H]
	\centering
	\footnotesize
	\caption{Validation and evidence signals used by \textsf{SKILL.nb}. Runtime never accesses benchmark labels. Calibration-split labels may be used only for offline threshold analysis; final evaluation labels are metrics-only.}
	\label{tab:validation_signals}
	\begin{tabularx}{\linewidth}{@{}l l l l X@{}}
		\toprule
		\textbf{Signal} & \textbf{Source} & \textbf{Runtime access} & \textbf{Use} & \textbf{Notes} \\
		\midrule
		Gate pass/fail & Browser state + task data & Yes & Runtime validation & Deterministic DOM, URL, visibility, text, form-value, and count checks \\
		Trace cluster $\mathcal{C}(q)$ & Retrieval + metadata & No & Create & Similarity may use learned or heuristic retrieval over prior logs \\
		Step evidence $n_i^{\mathrm{evidence}}$ & Distillation + validation & No & Formalize & Counts traces supporting step $s_i$ in the workflow version under consideration \\
		Repair count $m_i^{\mathrm{repair}}$ & Accepted repairs & No & Demote & Failed fallbacks do not count; repair must pass maintenance validation \\
		Repair burden $\rho_{\mathrm{repair}}$ & Accepted repair token costs & No & Retire & Workflow-lineage aggregate normalized by calibration repair cost \\
		Threshold violation & Offline validation & No & Feasibility & Candidate threshold admits an action exceeding $J_{\mathrm{perf}}^{\mathrm{ref}}+\epsilon$ \\
		Benchmark success & Evaluator & No & Metrics only & Final labels are not available to runtime or threshold estimation \\
		\bottomrule
	\end{tabularx}
\end{table}

\subsection{Runtime System Details}\label{app:runtime_system}

This section collects the operational details supporting \S\ref{ssec:runtime}. It gives the existing-workflow execution algorithm together with the fuller runtime semantics for gate-conditioned execution, drift recovery, experience distillation, and asynchronous maintenance.

\subsubsection{Execution Algorithms}\label{app:runtime_algorithm}

Runtime execution is decomposed into two substantive procedures. Algorithm~\ref{alg:runtime} manages run state, proposal accumulation, and recovery escalation, while Algorithm~\ref{alg:perform_step} encapsulates gate checking, local repair, and the realization cascade. Failure distillation and the recovery trigger are simple bookkeeping operations, so we describe them in prose rather than as separate algorithm floats.

\begin{algorithm}[!b]
	\footnotesize
	\caption{Top-level existing-workflow execution loop}
	\label{alg:runtime}
	\begin{algorithmic}[1]
		\Require query $q$, knowledge repo $\mathcal{K}$, runtime memory $\mathcal{M}$,
		execution log $\mathcal{R}_{\mathrm{logs}}$, policy parameter $\tau_{\mathrm{recover}}$, env $\mathcal{E}$
		\Ensure result $y$
		\State $\mathcal{W}_{v_L} \gets \mathrm{Retrieve}(q, \mathcal{K})$; $S \gets \mathcal{W}_{v_L}.S$
		\State $\mathcal{I}_{\mathrm{aff}} \gets \varnothing$; $\mathrm{patches} \gets \varnothing$; $\mathcal{B}_{\mathrm{mem}} \gets \varnothing$
		\For{$i = 1$ \textbf{to} $|S|$}
		\State $\mathcal{N}_i \gets \mathrm{RetrieveMem}(q, x_t, S[i], \mathcal{M})$
		\State $(\mathrm{ok}, p_i, \rho_i, r_i) \gets \Call{PerformStep}{\mathcal{W}_{v_L}, i, \mathcal{N}_i, \mathcal{E}}$
		\If{\textbf{not} $\mathrm{ok}$}
		\State distill failure memory $\rho_f$ and proposal $\Pi_f$ from $(q, i, x_t, \mathcal{R}_{\mathrm{logs}})$
		\State add non-empty $\rho_f$ to $\mathcal{B}_{\mathrm{mem}}$; submit non-empty $\Pi_f$
		\State $\mathcal{I}_{\mathrm{aff}} \gets \mathcal{I}_{\mathrm{aff}} \cup \{i\}$
		\Else
		\State add non-empty $\rho_i$ to $\mathcal{B}_{\mathrm{mem}}$ and non-empty $p_i$ to $\mathrm{patches}$
		\If{$r_i$ \textbf{or} $p_i \neq \varnothing$} \State $\mathcal{I}_{\mathrm{aff}} \gets \mathcal{I}_{\mathrm{aff}} \cup \{i\}$ \EndIf
		\EndIf
		\State $\rho_{\mathrm{aff}} \gets |\mathcal{I}_{\mathrm{aff}}| / |S|$
		\If{$\rho_{\mathrm{aff}} \ge \tau_{\mathrm{recover}}$}
		\State $(p_g, y_g) \gets \mathrm{GlobalRecoverAndExecute}(\mathcal{W}_{v_L}, i, q, \mathcal{N}_i, \mathcal{R}_{\mathrm{logs}}, \mathcal{E}, \mathcal{K})$
		\If{$p_g \neq \varnothing$} \State $\mathrm{patches} \gets \mathrm{patches} \cup \{p_g\}$ \EndIf
		\If{$y_g \neq \mathrm{success}$}
		\If{$\mathcal{B}_{\mathrm{mem}} \neq \varnothing$} \State $\mathcal{M} \gets \mathrm{Consolidate}(\mathcal{M}, \mathcal{B}_{\mathrm{mem}})$ \EndIf
		\State \Return $\mathrm{failure}$
		\EndIf
		\State \textbf{break}
		\ElsIf{\textbf{not} $\mathrm{ok}$}
		\If{$\mathcal{B}_{\mathrm{mem}} \neq \varnothing$} \State $\mathcal{M} \gets \mathrm{Consolidate}(\mathcal{M}, \mathcal{B}_{\mathrm{mem}})$ \EndIf
		\State \Return $\mathrm{failure}$
		\EndIf
		\EndFor
		\If{$\mathrm{patches} \neq \varnothing$}
		\State $\mathcal{W}' \gets \mathrm{BuildProposal}(\mathcal{W}_{v_L}, \mathrm{patches}, \mathcal{I}_{\mathrm{aff}})$
		\State $\mathrm{Submit}(\mathcal{W}')$
		\EndIf
		\If{$\mathcal{B}_{\mathrm{mem}} \neq \varnothing$} \State $\mathcal{M} \gets \mathrm{Consolidate}(\mathcal{M}, \mathcal{B}_{\mathrm{mem}})$ \EndIf
		\State \Return $\mathrm{success}$
	\end{algorithmic}
\end{algorithm}

\begin{algorithm}[H]
	\footnotesize
	\caption{Step execution with gate checking, local repair, and realization fallback}
	\label{alg:perform_step}
	\begin{algorithmic}[1]
		\Require workflow $\mathcal{W}_{v_L}$, step index $i$, retrieved memory $\mathcal{N}_i$, env $\mathcal{E}$
		\Ensure $(\mathrm{ok}, p_i, \rho_i, r_i)$
		\State $p_i \gets \varnothing$; $\rho_i \gets \varnothing$; $r_i \gets \mathrm{false}$
		\If{\textbf{not} $\gamma_{i,\mathrm{pre}}(x_t)$}
		\State diagnose the mismatch between expected assumptions and current state
		\If{the mismatch is attributable to environmental drift and a local repair succeeds}
		\State $r_i \gets \mathrm{true}$; record any induced patch $p_i$
		\Else
		\State \Return $(\mathrm{false}, p_i, \rho_i, r_i)$
		\EndIf
		\EndIf
		\State attempt realizations in order $C_i \to P_i \to I_i$ until one satisfies the postcondition gate
		\If{the attempted realization is accepted}
		\State distill any reusable local-repair memory into $\rho_i$
		\State \Return $(\mathrm{true}, p_i, \rho_i, r_i)$
		\EndIf
		\State \Return $(\mathrm{false}, p_i, \rho_i, r_i)$
	\end{algorithmic}
\end{algorithm}

Algorithm~\ref{alg:runtime} is the algorithmic realization of the runtime loop in \S\ref{ssec:runtime}. It manages run-level state: retrieving the released workflow, maintaining the affected-step set $\mathcal{I}_{\mathrm{aff}}$, accumulating local patches, and deciding when to terminate, recover globally, or submit a workflow-level proposal.

Algorithm~\ref{alg:perform_step} encapsulates gate-conditioned step execution. It checks the precondition gate, attempts local repair when the mismatch appears to be environmental drift, then attempts the code $\to$ procedure $\to$ intent cascade until one realization satisfies the postcondition. Its outputs are the success flag $\mathrm{ok}$, a local patch $p_i$, a reusable memory item $\rho_i$, and the repair indicator $r_i$.

When a step fails, runtime extracts a failure-sourced memory item $\rho_f$ and a workflow-level proposal $\Pi_f$ from the current execution trace. The memory item may later be consolidated into $\mathcal{M}$, while the proposal is submitted for offline maintenance review. Recovery escalation uses the normalized affected-step ratio $\rho_{\mathrm{aff}} = |\mathcal{I}_{\mathrm{aff}}|/|S|$ and triggers global recovery when $\rho_{\mathrm{aff}} \ge \tau_{\mathrm{recover}}$; the next subsection gives the full criterion.

\subsubsection{Gate-Conditioned Execution}\label{app:runtime_details}

Runtime handles a query $q$ in either \emph{existing-workflow} mode or \emph{no-workflow} mode. In existing-workflow mode, the latest released workflow is the sole authoritative runtime artifact. In no-workflow mode, the agent synthesizes a provisional workflow $\hat{\mathcal{W}}$ from the task intent and current observable state; similar released workflows, supporting artifacts, and same-site or same-domain logs may be used only as advisory context. Gates for either released or provisional steps are executable predicates over browser-observable state, and cannot query benchmark evaluators, final task labels, or hidden oracle state.

During either mode, runtime maintains a temporary per-run memory $\mathcal{M}$ alongside the authoritative repository $\mathcal{K}$. This memory is mutable but non-authoritative: it may store transient observations, local repair traces, and provisional routines for the current run, but it cannot update $\mathcal{K}$ directly. Candidate updates derived from $\mathcal{M}$ become durable only after offline review and promotion.

\subsubsection{Runtime Recovery Criterion}\label{app:recovery_details}

During execution, runtime maintains a run-level instability signal to determine when local repair is no longer sufficient. A step is marked as affected when it fails locally, requires an accepted local repair, or produces a non-empty step-local patch $p_i$ indicating drift. Let
\[
	\mathcal{I}_{\mathrm{aff}} = \{ i \mid \text{step } i \text{ has shown instability in the current run} \}
\]
denote the set of unique affected steps in the current run. Each step contributes at most once, so repeated local difficulty on the same step does not artificially inflate the run-level signal. The normalized affected-step ratio is
\[
	\rho_{\mathrm{aff}} = \frac{|\mathcal{I}_{\mathrm{aff}}|}{|S|},
\]
where $|S|$ is the number of steps in the current workflow.

Runtime escalates to workflow-level recovery when this ratio crosses the runtime recovery control $\tau_{\mathrm{recover}}$:
\[
	e_i =
	(\rho_{\mathrm{aff}} \ge \tau_{\mathrm{recover}}).
\]
This normalization makes the recovery control comparable across workflows of different lengths. When $e_i$ holds, the agent invokes a global recovery routine that re-plans and executes the remaining steps using retrieved memories, execution history, and prior workflow versions as context~\citep{ouyang2025reasoningbank}. If recovery cannot make progress under the available task context and permissions, the run terminates as a failure. When progress requires an external precondition unavailable to the agent, such as credentials or user approval, runtime pauses for operator input or terminates. This fallback is outside the autonomous lifecycle policy and is not counted as a method capability in evaluation.

\subsubsection{Experience Distillation}\label{app:distillation_details}

Runtime traces are distilled only into non-authoritative evidence in $\mathcal{M}$ or into candidate workflow-level proposals for offline review. These outputs may summarize reusable procedural fragments, local repairs, missing preconditions, guardrails, or drift signatures, but raw experiences never enter $\mathcal{K}$ directly; durable updates require offline verification and promotion.

\subsubsection{Asynchronous Maintenance}\label{app:maintenance_details}

Once runtime agents submit a candidate $\mathcal{W}'$, offline maintenance agents verify, refactor, and promote validated workflows into $\mathcal{K}$, with their inference calls counted in maintenance cost. They also review the lessons and repair-derived workflow changes distilled from the run's temporary memory for possible promotion into supporting repository artifacts.

Raw runtime experience is first distilled into non-authoritative per-run evidence in $\mathcal{M}$, and only then do candidate workflow changes or supporting artifacts enter gated promotion. This progression lets \textsf{SKILL.nb} preserve the adaptability of learned agent experience while keeping durable notebook updates and reusable assets under code-level governance.

\noindent\textbf{Verification and Promotion.}
Upon receiving a proposal $\mathcal{W}'$ or failure-sourced proposal $\Pi_f$, maintenance agents refine the candidate against execution traces, stripping exploratory detours and backtracking actions to yield a minimal, reproducible artifact. They resolve duplicates and conflicts against $\mathcal{K}$, then verify the candidate using logged traces, cached regression checks, or side-effect-controlled reruns where available. This consolidation is deliberately stricter than append-only memory: candidate changes must correspond to concrete workflow edits and pass deterministic gate checks before promotion. Gate definitions $\Gamma_i$ are updated where needed to reflect observed drift. Each promotion creates a new version and release in $\mathcal{K}$ and triggers downstream maintenance jobs.

\subsection{Adaptive Thresholds: Technical Details}\label{app:adaptive_thresholds}

This section records the deployed threshold-estimation procedure supporting \S\ref{ssec:adaptive_thresholds}. The intent is reproducibility rather than a distribution-free guarantee: thresholds are filtered by replay on logged decision opportunities, then shrunk toward pooled behavior when group evidence is sparse.

\noindent\textbf{Regularized interpretation.} Viewed more formally, the pooled-plus-group procedure in \S\ref{ssec:adaptive_thresholds} can be read as an empirical-Bayes-style shrinkage heuristic. Under a local-quadratic approximation to the within-group costs, and when the pooled cost dominates the regularization pull, the operational procedure is well approximated by a conservative single-pass coordinate-descent update to the following objective:
\begin{equation}\label{eq:adaptive_erm}
	\min_{\{\tau_d^g\},\, \{\tau_d^{\mathrm{pool}}\}}
	\sum_{d}
	\left[
	\hat{C}^{(d)}_{\mathrm{maint}}(\mathrm{pool}, \tau_d^{\mathrm{pool}})
	+
	\sum_{g \in \mathcal{G}_d}
	\Bigl(
	\hat{C}^{(d)}_{\mathrm{maint}}(g, \tau_d^g)
	\;+\;
	\lambda_{g,d}\,\bigl(\tau_d^g - \tau_d^{\mathrm{pool}}\bigr)^2
	\Bigr)
	\right]
\end{equation}
subject to $\mathrm{WilsonUCB}_{1-\alpha}[\hat{V}^{(d)}(\mathrm{pool}, \tau_d^{\mathrm{pool}})] \le V_{\max}^{(d)}$ and $\mathrm{WilsonUCB}_{1-\alpha}[\hat{V}^{(d)}(g, \tau_d^g)] \le V_{\max}^{(d)}$ for all $d$ and $g \in \mathcal{G}_d$, with $\tau_d^g, \tau_d^{\mathrm{pool}} \in \mathcal{T}_d$. Calibrating $\lambda_{g,d} \propto 1/n_{g,d}$ yields the shrinkage weight $\omega_{g,d} = n_{g,d}/(n_{g,d}+n_0)$, where $n_0$ is a reference sample size controlling the transition from pooled to group-specialized behavior, and the unconstrained blend $\omega_{g,d}\,\hat{\tau}_d^{g} + (1-\omega_{g,d})\,\hat{\tau}_d^{\mathrm{pool}}$. This blend has the usual partial-pooling behavior~\citep{gelman2013bayesian}: small groups are shrunk toward the pooled default, while well-supported groups remain close to their own empirical optimum. Because shrinkage is continuous in $n_{g,d}$, no separate minimum-sample specialization gate is needed: sparse groups remain near pooled behavior automatically. The runtime rule substitutes the constrained sweep output $\hat{\tau}_d^{g}$ for the unconstrained minimizer when the group has a nonempty feasible set. When the unconstrained minimum is infeasible but $\mathcal{F}_{g,d}$ is nonempty, this substitution is conservative because it shrinks from a feasible point. Section~\ref{app:decoupling} gives the deployed estimation procedure.

\subsubsection{Threshold Estimation Procedure}\label{app:group_assignment}\label{app:decoupling}

Before threshold estimation, offline maintenance agents assign each workflow and step to a group using the versioned metadata maps stored with the artifact. These group descriptors are maintenance-produced artifact metadata, not a fixed benchmark taxonomy. They may change when a workflow version is updated, repaired, or re-reviewed. Descriptors are selected from a constrained schema and canonicalized before threshold estimation; the schema and canonicalization rules are frozen for each calibration run. For threshold calibration, group assignment is performed chronologically from metadata available at the time of the logged decision. Later outcomes, benchmark labels, and final task success are excluded from the grouping input. Rare or novel descriptors back off to coarser parents, and the exact group assignment used for each decision opportunity is logged with the workflow version. Descriptors are restricted to reusable, observable properties such as site family, task type, action type, step type, and interface properties such as whether a form is dynamic. They do not include benchmark task identifiers, hidden evaluator outputs, final answers, or success labels. The resulting labels determine which logged decision opportunities contribute to group-specific estimates.

The retirement normalizer is fixed from calibration logs as $c_{\mathrm{ref}}=\max_{e \in \mathcal{R}_{\mathrm{cal}}} c(e)$, the maximum accepted repair-event token cost observed during calibration. If $\mathcal{R}_{\mathrm{cal}}$ is empty, automatic retirement by repair burden is deferred to maintenance review. For each decision $d$, the candidate set $\mathcal{T}_d$ is the sorted set of unique replay values of that decision's signal, not a hand-tuned grid. For each $\tau \in \mathcal{T}_d$, replay estimates maintenance cost $\hat{C}^{(d)}_{\mathrm{maint}}(g,\tau)$ and violation rate $\hat{V}^{(d)}(g,\tau)$ on threshold-estimation cases $\mathcal{D}_{g,d}$. A case is used only when the threshold-estimation log contains the candidate action with its validation outcome and maintenance-token cost. The sweep keeps candidates whose pointwise one-sided Wilson upper bound is at most $V_{\max}^{(d)}$ and selects the feasible candidate with lowest replay-estimated maintenance cost. The same sweep is run on pooled data and on each group with usable support.

For $k=k_{g,d}(\tau)$, $n=n_{g,d}$, $\hat{p}=k/n$, and $z=\Phi^{-1}(1-\alpha)$, the one-sided Wilson upper bound is
\[
	\mathrm{WilsonUCB}_{1-\alpha}(k,n)
	=
	\frac{
		\hat{p} + z^2/(2n)
		+ z\sqrt{\hat{p}(1-\hat{p})/n + z^2/(4n^2)}
	}{
		1 + z^2/n
	}.
\]
A threshold is feasible when this upper bound is at most $V_{\max}^{(d)}$. Because the final threshold is selected after sweeping over $\mathcal{T}_d$, this bound is used as a conservative pointwise filter rather than a formal uniform confidence guarantee after adaptive selection.

The deployed rule uses the sample-size shrinkage weight $\omega_{g,d}=n_{g,d}/(n_{g,d}+n_0)$ from \S\ref{ssec:adaptive_thresholds}. This is the usual partial-pooling form: groups with little replay support stay near the pooled threshold, while well-observed groups move toward their own replay estimate. The implementation is parameterized directly by $n_0$ and does not estimate a curvature term. After shrinkage, the blended threshold may not be one of the replay-feasible candidate values, so the deployed rule projects it back onto the finite feasible set:
\[
	\Pi_{\mathcal{F}}(x) = \arg\min_{\tau \in \mathcal{F}} |\tau - x|.
\]
Ties are broken toward $\hat{\tau}_d^{\mathrm{pool}}$. If $n_{g,d}=0$, the method skips the specialized branch and uses the pooled branch when $\mathcal{F}^{\mathrm{pool}}_d$ is nonempty. If $n_{g,d}>0$ but $\mathcal{F}_{g,d}=\emptyset$, the method does not automatically apply the pooled threshold to that group. It routes the corresponding lifecycle action to maintenance review, because group-specific replay found no threshold satisfying the violation budget. If the pooled feasible set is also empty, the method returns $\operatorname{defer}_d$. This value is not a numeric threshold; it suppresses automatic thresholded action for decision $d$ until maintenance review. This avoids applying a threshold for which replay found no candidate satisfying the violation budget.

Runtime outcomes feed the threshold signals as follows: successful executions add workflow-level evidence for creation and step-level evidence for formalization. Accepted local repairs add a count signal for demotion and a token-weighted burden signal for retirement. The same threshold mechanism applies to all four lifecycle decisions: \texttt{create} releases a sufficiently supported workflow, \texttt{form} adds executable realizations to stable steps, \texttt{demote} returns repaired or unstable steps to NL-guided execution, and \texttt{retire} removes a workflow when normalized repair burden is widespread.

\begin{table}[H]
	\centering
	\small
	\caption{Lifecycle decisions governed by replay-calibrated thresholds.}
	\label{tab:lifecycle_decisions}
	\begin{tabularx}{\linewidth}{@{}l l l l X@{}}
		\toprule
		Decision & Signal & Threshold & Action & Required validation \\
		\midrule
		Create & $|\mathcal{C}(q)|$ & Count & provisional $\rightarrow$ released & $\mathrm{ValidW}_v$ \\
		Form & $n_i^{\mathrm{evidence}}$ & Count & add/use $C_{i,v}$ & $\mathrm{ValidC}_{i,v}$ \\
		Demote & $m_i^{\mathrm{repair}}$ & Count & executable $\rightarrow$ NL-guided successor & accepted repairs are gate- and validation-approved \\
		Retire & $\rho_{\mathrm{repair}}(\mathcal{W}_v)$ & Burden & released $\rightarrow$ retired & $\mathrm{RetireOK}_v$ after maintenance review \\
		\bottomrule
	\end{tabularx}
\end{table}

\subsubsection{Replay-Relative Scope of the Feasibility Check}\label{app:replay_compliance}

Projection does not claim the threshold is safe under future shifts. It only ensures that the deployed value is one of the candidates that passed the replay filter on the logged opportunities used for estimation. The UCB check is replay-relative: it filters thresholds using $\hat{V}^{(d)}$, the violation rate estimated by changing one candidate threshold at historical decision points while keeping the rest of the logged repository trajectory fixed. It does not control the true violation rate under arbitrary future workload or interface shift, and it does not model second-order effects where a different threshold would have changed later repository contents, group assignments, or repair opportunities. For this reason, the paper describes the result as UCB-bounded replay feasibility rather than a deployment-time safety guarantee.

\subsubsection{Why Threshold Control Rather Than End-to-End RL}\label{app:rlvr_threshold_comparison}

Table~\ref{tab:rlvr_vs_thresholds} summarizes why \textsf{SKILL.nb} uses thresholded lifecycle control rather than end-to-end RL or RLVR. The distinction is not that RLVR is unsuitable in general. Rather, the policy target here is a durable repository of workflow artifacts, and the available counterfactual evidence is local to logged lifecycle decisions.

\begin{table}[H]
	\centering
	\small
	\caption{Why \textsf{SKILL.nb} uses thresholded lifecycle control rather than end-to-end RLVR. The policy governs durable workflow artifacts, not token generation or primitive browser actions.}
	\label{tab:rlvr_vs_thresholds}
	\begin{tabularx}{\linewidth}{@{}lXX@{}}
		\toprule
		\textbf{Property} & \textbf{End-to-end RL/RLVR} & \textbf{\textsf{SKILL.nb} threshold control} \\
		\midrule
		Policy target & Model or browser-action policy & Repository lifecycle policy \\
		Action granularity & Token or primitive action level & Release, formalize, demote, retire \\
		Feedback signal & Task-level verifiable reward & Gates, accepted repairs, and replay validation \\
		Policy search & Online exploration or broad offline optimization & Constrained replay-calibrated threshold selection \\
		Counterfactual burden & Long-horizon trajectory effects & Logged lifecycle decision opportunities \\
		Safety scope & Depends on verifier and exploration regime & Replay-relative UCB filtering plus runtime gates \\
		Interpretability & Learned policy may be opaque & Auditable thresholds by group and decision \\
		\bottomrule
	\end{tabularx}
\end{table}

\subsection{Limitations and Scope}\label{app:limitations_scope}

The adaptive-threshold procedure has bounded scope in several important ways. Its UCB-bounded filtering is replay-relative through the estimator $\hat{V}^{(d)}$, rather than a blanket deployment-time guarantee under arbitrary distribution shift. Replay changes one candidate threshold at a time while holding the rest of the logged lifecycle fixed, so it does not model second-order effects on future repository contents, future group assignments, or later repair opportunities. The offline routine is therefore a practical single-pass approximation to the joint four-decision problem rather than a fully joint optimizer. We also do not evaluate alternative threshold estimators such as Bayesian optimization over the empirical replay breakpoints or learned predictors from group features. Direct offline partial pooling with UCB-constrained sweeps is chosen here for sample efficiency on finite replay candidate sets, interpretability, and explicit feasibility filtering. Appendix~\ref{app:adaptive_thresholds} provides the technical details.

The method also depends on the quality of the gates and metadata used to produce these signals. Evidence and repair signals are only as reliable as the gate specifications $(\gamma_{i,\mathrm{pre}}, \gamma_{i,\mathrm{post}})$ and the offline metadata tags used to group workflows and steps. Gate false positives can admit an invalid step execution or repair, while false negatives can trigger unnecessary fallback, demotion, or repair. Poor gates or miscalibrated tags can therefore distort the signals driving the lifecycle decisions. Separately, the benefits of specialization depend on recurring workload structure. This assumption is plausible in deployment settings such as help-desk or enterprise automation, where many requests recur as variations of the same underlying workflows, but it may fail in more heterogeneous environments. If few groups accumulate enough support, shrinkage keeps their deployed thresholds close to the pooled branch and the method behaves more like a pooled retuning baseline than a strongly specialized policy. The maintenance-cost proxy $c(e)=\mathrm{tok}_{\mathrm{in}}(e)+\mathrm{tok}_{\mathrm{out}}(e)$ measures LLM inference tokens only. It does not optimize for wall-clock latency, human handoff cost, storage overhead for the repository and logs, or developer review effort. For retirement, token-weighted repair burden is a maintenance-effort proxy rather than a complete measure of semantic criticality, so a cheap but critical repair can still be underweighted.

The empirical evaluation also has bounded scope. Benchmark evaluations provide snapshots of performance under fixed benchmark conditions, while the method's distinctive correction capabilities, including demotion, retirement, and re-formation after drift, operate over longer deployment timescales that benchmark runs do not naturally exercise. As a result, the paper characterizes the growth phase of workflow creation and step formalization more thoroughly than the correction phase.

\noindent\textbf{Broader impacts.}
More auditable workflow artifacts can make web agents easier to inspect, reproduce, and roll back when they fail. This may reduce silent regressions in benign automation settings such as internal support or software-maintenance workflows. The same capability could also make undesired web automation more reliable, or preserve traces that contain sensitive visual state or credentials if deployed without redaction. For this reason, our claims are limited to controlled benchmarks, and deployment should pair lifecycle governance with credential scoping, trace redaction, rate limits, and task-level authorization controls.

\section{Experimental Details}\label{app:experimental_details}

This appendix records protocol details for the main experiments in \S\ref{sec:experiments}. The main text summarizes the benchmark choices, shared harness, and statistical conventions, while this appendix pins the run configuration, persistent-state rules, re-execution protocol, repair protocol, and confidence-interval calculations. Additional component-removal and transfer diagnostics appear in Appendix~\ref{app:additional_experiments}.

\subsection{Evaluation Protocol}\label{app:evaluation_protocol}

All reported experiments use identical infrastructure and a fixed execution configuration. All compared methods are run with the same GPT~5.3 Codex model, tool access, and execution budget in our evaluation harness for a fair comparison, so the main-text tables and figures report a unified re-evaluation rather than numbers copied directly from the original papers. For baseline methods, the shared harness wraps the released public implementations and preserves their native persistent-state and update paths rather than replacing them with a common re-implementation. On WebArena-Verified, the benchmark table in the main text reports a fresh-start full benchmark round in which each method starts without persistent learned state and builds any repository or memory online during that round. Separately, the lifecycle protocol runs for five rounds over the full 812-task benchmark. For persistent-state methods, the GitLab version-drift evaluation includes fresh-start runs on GitLab~15.7, GitLab~16.11, and GitLab~18.9, plus old-to-target conditions that build old-version state on GitLab~15.7 and reuse restored snapshots on GitLab~16.11 and GitLab~18.9. We choose GitLab~16.11 as an intermediate target because documented 16.x interface changes overlap with GitLab task families in WebArena-Verified, and use GitLab~18.9 as a longer-horizon target. Appendix~\ref{app:gitlab-version-drift} gives the detailed GitLab protocol and version-selection rationale. The benchmark environment is reset for each task execution. What persists varies by method: AWM\textsubscript{online} carries its induced workflow memory, ReasoningBank carries its distilled reasoning bank, and \textsf{SKILL.nb} carries its workflow repository $\mathcal{K}$, associated workflow state, event logs, and lifecycle-policy state. CodeAct has no persistent learned state or released-artifact maintenance loop and is therefore omitted from the lifecycle figures and GitLab old-state reuse evaluation. We do not synchronize update frequencies across methods. Each persistent-state method updates its state according to its own design during the multi-round protocol.

\noindent\textbf{Task order and online state.}
Each WebArena-Verified round uses a shuffled task order, and the same order for that round is applied to all methods. The order is reshuffled between rounds. After each task, transient agent context is cleared; persistent method state is not reset between tasks or rounds. We do not report multiple independent order seeds in this draft, so results should be interpreted as performance under the shared online-learning orders used here rather than as order-invariant estimates. Persistent state is allowed to store reusable task-family structure, natural-language procedures, executable workflow components, gates, and repair history. It is not allowed to store hidden evaluator outputs, benchmark success labels, final answers, or benchmark-specific task identifiers as reusable workflow fields. URLs may appear only when they are part of the observable task state or a reusable site-level navigation pattern.

\noindent\textbf{Lifecycle perturbations.}
For each lifecycle round, we generate one perturbation set offline and apply it to every method. Eligible tasks receive a same-site start URL sampled by random walk; tasks that require a benchmark-specified non-default start URL keep that required condition. Language perturbations are generated at the WebArena-Verified \texttt{intent\_template} level, e.g., ``Get name(s) of reviewer(s) who mention \{\{description\}\} for the product on the current page,'' rather than by rewriting fully instantiated tasks. An LLM proposes paraphrases while preserving slot variables, intent, and task parameters. We filter candidates with a sentence encoder, require semantic similarity above 90\%, and choose paraphrases with a penalty on word-level similarity so the maximum planned number of rounds uses diverse but semantically close wording. The paraphrase bank is generated once before evaluation and then reused across methods.

The exact template-paraphrasing prompt is:
\begin{quote}\small
You are paraphrasing WebArena-Verified intent templates for controlled evaluation. Rewrite the template in different wording while preserving the exact task intent, all required outputs, and every slot variable exactly as written. Do not add constraints, remove constraints, change entities, change the requested output type, or alter the success condition. Keep all \{\{slot\}\} placeholders unchanged. Return $N$ paraphrases, one per line.
\end{quote}

For \textsf{SKILL.nb}, the experimental hyperparameters are fixed unless otherwise noted: $V_{\max}^{(d)}=0.05$ for all lifecycle decisions, $\alpha=0.05$, $n_0=2$, and $\tau_{\mathrm{recover}}=0.25$. These values define the replay-feasibility budget, UCB tail probability, shrinkage reference sample size, and runtime recovery trigger used throughout the experiments. The \textsf{SKILL.nb} runtime and maintenance agents are implemented using \texttt{opencode}.

\noindent\textbf{Reproducibility record.}
Table~\ref{tab:reproducibility_record} records the fixed run configuration used for the reported experiments.

\begin{table}[t]
	\centering
	\small
	\caption{Reproducibility record for the reported experiments. For baseline methods marked with $^\ast$, temperature settings follow the corresponding original work where specified. All methods are re-evaluated with \texttt{gpt-5.3-codex}, and baseline runs use \texttt{high} reasoning effort for fair comparison.}\label{tab:reproducibility_record}
	\begin{tabular}{@{}l p{0.68\linewidth}@{}}
		\toprule
		Field & Value \\
		\midrule
		\multicolumn{2}{@{}l@{}}{\textbf{Methods}} \\
		\midrule
		\multicolumn{2}{@{}l@{}}{\textbf{CodeAct$^\ast$}} \\
		Model & \texttt{gpt-5.3-codex} \\
		Reasoning effort & \texttt{high} \\
		Temperature & $0.6$ (not reported in the original paper, so we use the value matching \textsf{SKILL.nb}) \\
		\midrule
		\multicolumn{2}{@{}l@{}}{\textbf{AWM\textsubscript{online}$^\ast$}} \\
		Model & \texttt{gpt-5.3-codex} \\
		Reasoning effort & \texttt{high} \\
		Temperature & $0.0$ for action generation, workflow induction, and memory update \\
		\midrule
		\multicolumn{2}{@{}l@{}}{\textbf{ReasoningBank$^\ast$}} \\
		Model & \texttt{gpt-5.3-codex} \\
		Reasoning effort & \texttt{high} \\
		Temperature & $0.7$ for generation, $1.0$ for memory extraction, $0.0$ for classification \\
		\midrule
		\multicolumn{2}{@{}l@{}}{\textbf{\textsf{SKILL.nb}}} \\
		Model & \texttt{gpt-5.3-codex} \\
		Reasoning effort & \texttt{medium} for execution, \texttt{high} for repair and drift handling \\
		Temperature & $0.2$ for execution, creation, and formalization, $0.6$ for repair and drift handling \\
		Lifecycle thresholds & $V_{\max}^{(d)}=0.05$, $\alpha=0.05$, $n_0=2$, $\tau_{\mathrm{recover}}=0.25$ \\
		Agent implementation & \texttt{opencode}\tablefootnote{\url{https://github.com/opencode-ai/opencode}} \\
		\midrule
		\multicolumn{2}{@{}l@{}}{\textbf{Benchmarks and evaluators}} \\
		\midrule
		WebArena-Verified &
		Commit \texttt{6473f72db5dcefc97b5725b59e734504edc28a21}\tablefootnote{\url{https://github.com/ServiceNow/webarena-verified}} \\
		WebArena-Verified hard subset & $258$ tasks for component and threshold diagnostics \\
		GitLab single-site subset & $180$ WebArena-Verified tasks from the full dataset supporting the main controlled version-drift evaluation \\
		Mind2Web &
		Commit \texttt{33bd95caeee7bba22dd08ecc935845e15c5e5dc7}\tablefootnote{\url{https://github.com/OSU-NLP-Group/Mind2Web}} \\
		\midrule
		\multicolumn{2}{@{}l@{}}{\textbf{Execution protocol}} \\
		\midrule
		Max steps & $30$ per task \\
		Browser environment & Playwright MCP-controlled browser\tablefootnote{\url{https://github.com/microsoft/playwright-mcp}} with per-task environment reset \\
		\bottomrule
	\end{tabular}
\end{table}

For \textsf{SKILL.nb}, the persistent state specifically includes the workflow repository $\mathcal{K}$, workflow states (including provisional, released, and retired artifacts), supporting event logs, and the learned lifecycle-policy state. Reuse consistency is measured before any update. For each method, the denominator is the set of tasks solved in the initial WebArena-Verified round. We freeze the method-specific learned state used for each such success: the released workflow for \textsf{SKILL.nb}, the associated workflow-memory state for AWM\textsubscript{online}, and the retrieved reasoning-bank state for ReasoningBank. Each frozen artifact is re-executed three times under independent environment resets, shared start-URL perturbations, and shared intent-template paraphrases. A task is counted as maintaining reuse consistency only if all three re-executions succeed without editing the artifact.

Only artifacts that later fail under re-execution enter the repair protocol. Each failing artifact receives up to three repair attempts through that method's native state-update mechanism. We do not impose a synthetic cross-method validation gate. For \textsf{SKILL.nb}, a candidate repair must pass promotion before replacing the pinned workflow version. For AWM\textsubscript{online}, repair attempts use its workflow-memory update path; for ReasoningBank, they use its memory extraction and reasoning-bank update path.

Recovery at budget $B$ is the fraction of failed artifacts restored to success within $B$ native update attempts. After each accepted update attempt, regression is evaluated on the cached-trace subset of tasks that were passing before the candidate update. This subset is method-specific: it is drawn from that method's previously passing tasks with reusable cached traces available at repair time and therefore varies by update event. Reuse consistency, recovery, and regression are therefore reported with method-conditional denominators, and the repair/regression comparison characterizes native update behavior rather than a matched validation-gate ablation.

On Mind2Web, we report only a single benchmark round in which each method starts without task-specific persistent state. Methods may accumulate workflows or memories online as allowed by their design, but no re-execution or repair is evaluated there. All repeated-execution and repair results in this paper are therefore specific to WebArena-Verified under the shared evaluation setup.

For statistical reporting on WebArena-Verified, individual method rates use Wilson 95\% confidence intervals. The main overall success-rate comparison between \textsf{SKILL.nb} and the next-best baseline is additionally evaluated with a two-sided continuity-corrected McNemar test on per-task success outcomes over the shared 812-task benchmark. We do not report analogous paired tests for every per-site subset in this draft, so those breakdowns should be interpreted as descriptive point estimates.

The round-2 dip for \textsf{SKILL.nb} in Figure~\ref{fig:lifecycle_sustained}(a) coincides with demotions of workflows created late in round~1. Because several task templates appear only once or twice, those workflows had limited evidence before the next perturbed round. We treat this as a protocol-specific diagnostic rather than a standalone claim about the method.

\subsection{Lifecycle Curve Confidence Intervals}\label{app:ci_tables}

Table~\ref{tab:lifecycle_ci} reports the 95\% Wilson confidence intervals for the per-round success rates shown in Figure~\ref{fig:lifecycle_sustained}(a). Each cell is computed over the full 812-task WebArena-Verified round for that method and round. The McNemar test reported in the main text is computed from paired per-task outcomes over the same 812 tasks, while the marginal Wilson intervals for the single-round WebArena-Verified benchmark are reported directly in Table~\ref{tab:exp_summary}.

\begin{table}[t]
	\centering
	\small
	\caption{Per-round WebArena-Verified success rates with 95\% Wilson confidence intervals for Figure~\ref{fig:lifecycle_sustained}(a). Each entry is reported as SR [lower, upper], in percent.}\label{tab:lifecycle_ci}
	\begin{tabularx}{\linewidth}{l *{5}{>{\centering\arraybackslash}X}}
		\toprule
		Method & Round 1 & Round 2 & Round 3 & Round 4 & Round 5 \\
		\midrule
		AWM\textsubscript{online} & 46.4 [43.0, 49.9] & 43.8 [40.5, 47.3] & 45.1 [41.7, 48.5] & 44.0 [40.6, 47.4] & 39.4 [36.1, 42.8] \\
		ReasoningBank & 49.8 [46.3, 53.2] & 49.4 [46.0, 52.8] & 48.2 [44.7, 51.6] & 45.6 [42.2, 49.0] & 45.2 [41.8, 48.6] \\
		\textsf{SKILL.nb} & 53.7 [50.3, 57.1] & 52.0 [48.5, 55.4] & 54.3 [50.9, 57.7] & 55.4 [52.0, 58.8] & 55.7 [52.2, 59.0] \\
		\bottomrule
	\end{tabularx}
\end{table}

\begin{table}[t]
	\centering
	\small
	\caption{Budget-2 recovery and regression rates with 95\% Wilson confidence intervals for Figure~\ref{fig:lifecycle_sustained}(c). Entries are reported as rate [lower, upper], in percent.}\label{tab:repair_ci}
	\begin{tabularx}{\linewidth}{l >{\centering\arraybackslash}X >{\centering\arraybackslash}X}
		\toprule
		Method & Recovery at budget 2 & Regression at budget 2 \\
		\midrule
		AWM\textsubscript{online} & 58.0 [48.8, 67.0] & 17.0 [12.8, 22.4] \\
		ReasoningBank & 63.0 [53.1, 72.2] & 15.0 [10.9, 20.3] \\
		\textsf{SKILL.nb} & 72.9 [63.3, 80.8] & 4.2 [2.4, 7.4] \\
		\bottomrule
	\end{tabularx}
\end{table}

\subsection{Lifecycle Maintenance Costs}\label{app:maintenance-cost}

\begin{figure}[H]
	\centering
	\includegraphics[width=0.7\textwidth]{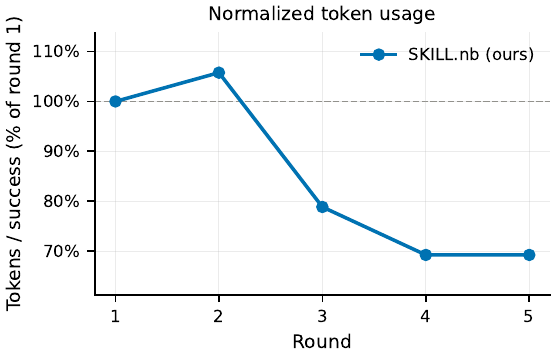}
	\caption{Maintenance token usage per successful task across five rounds, normalized to each method’s round-1 cost (100\%).}\label{fig:lifecycle-token-usage}
\end{figure}

Figure~\ref{fig:lifecycle-token-usage} reports the maintenance token usage per successful task across the five rounds of the lifecycle protocol. Because token counts are highly method-dependent, we report usage patterns normalized by each method's own round-1 maintenance cost (round~1 = 100\%), highlighting within-method cost dynamics. We observe that \textsf{SKILL.nb} experiences a steady decline in maintenance cost per task as its workflow repository $\mathcal{K}$ stabilizes, reaching 69.2\% of its initial round-1 cost by round~5. The small round-2 dip in success coincides with demotions of workflows created late in round~1, before enough evidence had accumulated to validate them reliably. We do not include other baselines in this analysis as they lack an explicit maintenance procedure that can be tracked and fairly measured against \textsf{SKILL.nb}.

\subsection{GitLab Version Drift}\label{app:gitlab-version-drift}

We use controlled GitLab version drift because persistent web-agent artifacts can fail when a maintained application changes outside the agent's control. This better matches how web-agent artifacts can fail in practice than synthetic UI perturbations chosen by the experimenter, since the user's underlying work remains similar while selectors, workflows, and stored assumptions may become stale. The experiment therefore tests reuse across an application-version change without treating it as evidence for broader drift settings.

This appendix provides protocol support for the main GitLab version-drift experiment in Section~\ref{ssec:gitlab-version-drift}. In Figure~\ref{fig:gitlab_drift}, v15, v16, and v18 abbreviate GitLab~15.7, GitLab~16.11, and GitLab~18.9, respectively. The candidate pool is the 180 single-site GitLab tasks from WebArena-Verified. We exclude multi-site tasks that include GitLab because those tasks add cross-site coordination and non-GitLab state, which would make the measured change less specific to GitLab application-version drift. The source deployment is GitLab~15.7. We evaluate GitLab~16.11 as an intermediate target and GitLab~18.9 as a longer-horizon target. We construct each target deployment by applying GitLab's official upgrade procedure\footnote{\url{https://docs.gitlab.com/update/upgrade/}} to the source deployment data, preserving the task data while changing the GitLab system version.

We selected these versions to test two levels of controlled drift. GitLab~15.7 matches the source environment for the benchmark tasks. GitLab~16.11 is the intermediate target because GitLab~16.0 introduced the new navigation experience and made the new Web IDE the default multi-file editor,\footnote{\url{https://docs.gitlab.com/releases/16/gitlab-16-0-released/}} while GitLab~16.11 redesigned the project overview page so project information and links appear in a sidebar.\footnote{\url{https://docs.gitlab.com/releases/16/gitlab-16-11-released/}} These changes overlap with the task mix: non-exclusive keyword grouping of the 180 single-site GitLab tasks gives about 40 issue-related tasks, 17 merge-request or review tasks, 20 file-editor or template tasks, and 132 project, repository, group, or member tasks. GitLab~18.9 is retained as a longer-horizon target after additional accumulated application changes.

\subsubsection{Example UI Drift in Project Navigation}

\begin{figure}[p]
	\centering
	\begin{minipage}{\linewidth}
		\centering
		\footnotesize\textbf{(a) GitLab~15.7}\par\vspace{0.25em}
		\includegraphics[width=0.9\linewidth]{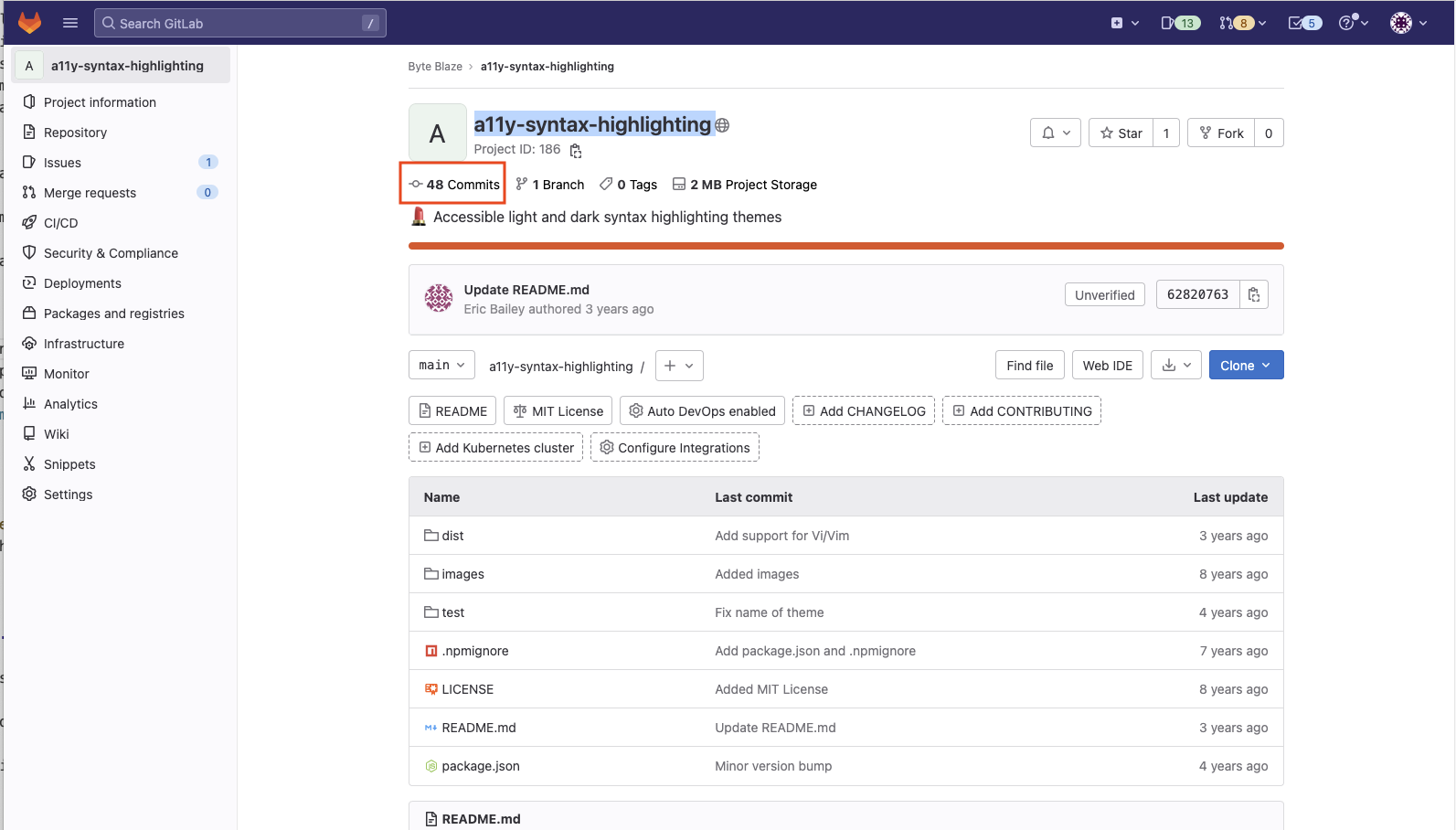}
	\end{minipage}\par\vspace{0.75em}
	\begin{minipage}{\linewidth}
		\centering
		\footnotesize\textbf{(b) GitLab~16.11}\par\vspace{0.25em}
		\includegraphics[width=0.9\linewidth]{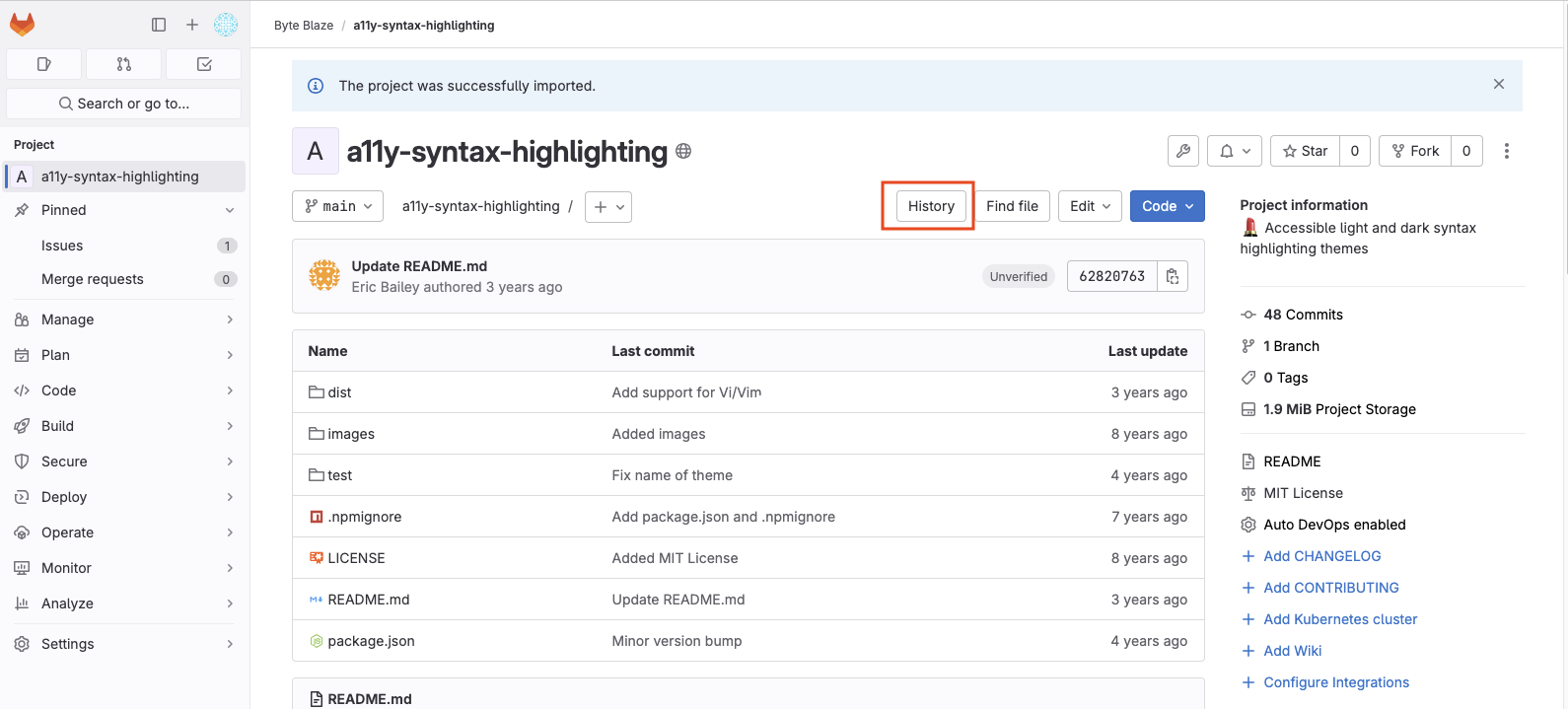}
	\end{minipage}\par\vspace{0.75em}
	\begin{minipage}{\linewidth}
		\centering
		\footnotesize\textbf{(c) GitLab~18.9}\par\vspace{0.25em}
		\includegraphics[width=0.9\linewidth]{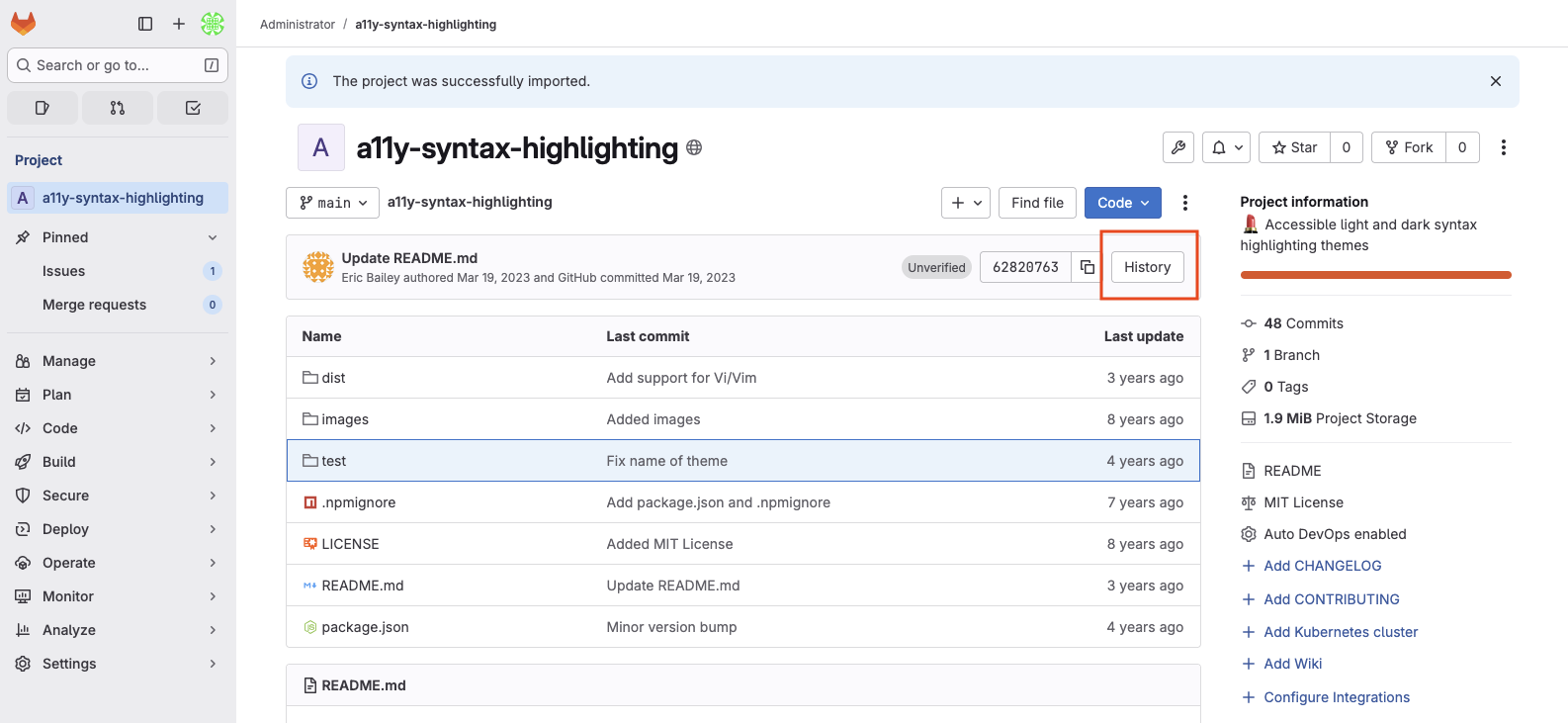}
	\end{minipage}
	\caption{Example project-level UI drift across GitLab versions. The screenshots show the same project view with the commit-history entry point highlighted in red. GitLab~15.7 exposes commit history through the commit-count link, while GitLab~16.11 and GitLab~18.9 expose a dedicated \texttt{History} button in different toolbar positions.}
	\label{fig:gitlab_ui_drift_example}
\end{figure}

Figure~\ref{fig:gitlab_ui_drift_example} shows the same project overview across the three GitLab deployments. The red boxes mark the UI control used to open commit history. In GitLab~15.7, there is no dedicated history button on the project file toolbar. The agent must instead click the commit-count link in the project summary. In GitLab~16.11 and GitLab~18.9, GitLab exposes a dedicated \texttt{History} button, but its position changes as the project summary and action toolbar are reorganized. This example illustrates why old procedural state can become stale even when the underlying task intent and project data are unchanged.

The underlying DOM changes as well. GitLab~15.7 exposes the entry point as a project-stat link, roughly \texttt{a.nav-link.stat-link}, with a nested \texttt{svg[data-testid=commit-icon]} and visible text \texttt{48 Commits}. GitLab~16.11 exposes a button-style anchor with visible text \texttt{History}, but without a dedicated test identifier. GitLab~18.9 adds \texttt{data-testid=last-commit-history} to the corresponding anchor. This supports the qualitative observation in Section~\ref{ssec:gitlab-version-drift}: newer GitLab versions can expose more stable DOM locators, and \texttt{data-testid} attributes are preferred by browser automation because they are less tied to layout or styling than class names.

\begin{table}[H]
	\centering
	\scriptsize
	\caption{DOM-level differences for the same commit-history navigation target across GitLab versions.}
	\label{tab:gitlab_history_dom_drift}
	\begin{tabularx}{\linewidth}{@{}lXX@{}}
		\toprule
		Version & Observed DOM signature & Automation implication \\
		\midrule
		GitLab~15.7 & \texttt{a.nav-link.stat-link} with visible text \texttt{48 Commits}; no dedicated history button & Requires selecting a project-stat link rather than a toolbar action \\
		GitLab~16.11 & \texttt{a.btn} with visible text \texttt{History}; no \texttt{data-testid} & Adds the history action, but selectors still depend on class names or visible text \\
		GitLab~18.9 & \texttt{data-testid=last-commit-history} on the \texttt{History} anchor & Provides a stable test identifier preferred for automated browser locators \\
		\bottomrule
	\end{tabularx}
\end{table}

We define \(\mathcal{D}_{\mathrm{mig}}\) as the subset of the 180 candidate tasks whose required task state and evaluator can be migrated or reconstructed on both target deployments. In the current migrated-run aggregate, \(|\mathcal{D}_{\mathrm{mig}}|/180 = 180/180\). All success rates in this experiment are computed on \(\mathcal{D}_{\mathrm{mig}}\).

We evaluate five conditions on the same \(\mathcal{D}_{\mathrm{mig}}\) denominator. The source fresh condition starts each method from empty persistent state on GitLab~15.7. The two target fresh conditions start each method from empty persistent state on GitLab~16.11 and GitLab~18.9. The two old-to-target reuse conditions first start each method from empty persistent state on GitLab~15.7 and run on \(\mathcal{D}_{\mathrm{mig}}\) to construct old-version state, then snapshot that state for target-version evaluation. For AWM\textsubscript{online}, the snapshot is the induced workflow memory. For ReasoningBank, the snapshot is the distilled reasoning bank, memory pool, and retrieval index as used by the implementation. For \textsf{SKILL.nb}, the snapshot includes the workflow repository \(\mathcal{K}\), workflow states, event logs needed by the lifecycle policy, and threshold and lifecycle state.

During target-version reuse evaluation, each target task restores the old-version snapshot or uses an isolated copy of that snapshot. Transient context, browser state, and environment state reset per task in all five conditions. Within-task fallback or local repair is allowed only when it is part of the method's normal execution. Any durable workflow, repository, memory, retrieval-index, event-log, threshold, or lifecycle update written during one GitLab~16.11 or GitLab~18.9 reuse task is discarded before later target tasks are scored. The fresh target-version conditions control for target-version difficulty, while the reuse conditions measure target-version success under old-state reuse rather than cumulative target-version relearning.

Table~\ref{tab:gitlab_version_drift_appendix} gives the appendix results. The fresh columns measure performance when the method starts empty on each version. The reuse columns measure target-version performance when initialized from the GitLab~15.7 persistent-state snapshot.

For each success-rate cell and each point in Figure~\ref{fig:gitlab_drift}, we compute a two-sided 95\% Wilson binomial confidence interval from the number of successful tasks \(k\) out of \(n=|\mathcal{D}_{\mathrm{mig}}|\) migrated tasks. With \(\hat{p}=k/n\) and \(z=1.96\), the interval is
\[
	\frac{\hat{p}+z^2/(2n) \pm z\sqrt{\hat{p}(1-\hat{p})/n+z^2/(4n^2)}}{1+z^2/n}.
\]
We multiply the endpoints by 100 when reporting percentages.

\begin{table}[H]
	\centering
	\scriptsize
	\setlength{\tabcolsep}{2.5pt}
	\caption{Appendix support table for the GitLab version-drift evaluation on \(\mathcal{D}_{\mathrm{mig}}\). Each persistent-state method is evaluated under source fresh start, target fresh starts, and target-version reuse from a GitLab~15.7 persistent-state snapshot. Cells report SR with 95\% Wilson confidence intervals.}\label{tab:gitlab_version_drift_appendix}
	\begin{tabular}{@{}lccccc@{}}
		\toprule
		Method                    & 15.7 fresh & 16.11 fresh & 15.7$\rightarrow$16.11 reuse & 18.9 fresh & 15.7$\rightarrow$18.9 reuse \\
		\midrule
		AWM\textsubscript{online} & 40.0 [33.1,47.3] & 48.3 [41.1,55.6] & 33.9 [27.4,41.1] & 48.9 [41.7,56.1] & 34.4 [27.9,41.6] \\
		ReasoningBank             & 45.6 [38.4,52.8] & 51.7 [44.4,58.9] & 41.1 [34.2,48.4] & 50.0 [42.8,57.2] & 38.9 [32.1,46.2] \\
		\textsf{SKILL.nb}         & 54.4 [47.2,61.6] & 61.7 [54.4,68.5] & 60.0 [52.7,66.9] & 61.1 [53.8,67.9] & 61.7 [54.4,68.5] \\
		\bottomrule
	\end{tabular}
\end{table}

\clearpage
\section{Additional Experiments}\label{app:additional_experiments}

\subsection{Controlled Component Ablations}\label{app:component_ablations}

In this section, we include a support ablation to address whether the final \textsf{SKILL.nb} behavior depends on its main components rather than on a single incidental design choice. On the 258-task WebArena-Verified hard subset, the experiment isolates selective code formalization with \(\text{\textsf{SKILL.nb}}_{\text{NL-only}}\), which keeps only workflow intents and natural-language procedures; fallback with \(\text{\textsf{SKILL.nb}}_{\text{code-only}}\), which disables fallback when executable code is unavailable or brittle; runtime validation with \(\text{\textsf{SKILL.nb}}_{\text{no-gates}}\), which removes precondition and postcondition checks; and lifecycle cleanup with \(\text{\textsf{SKILL.nb}}_{\text{no-demote}}\), which disables demotion and retirement. All variants use the same model, harness, task stream, repository schema, evaluation budget, and online maintenance setting; each variant maintains its own repository state. Threshold specialization is evaluated separately in Appendix~\ref{app:threshold-ablation}.

Table~\ref{tab:component_ablation} reports the resulting success, \textsf{SKILL.nb}-internal token cost, fallback, repair, and regression diagnostics.

\begin{table}[H]
	\centering
	\footnotesize
	\setlength{\tabcolsep}{3.2pt}
	\caption{Controlled component ablations on the 258-task WebArena-Verified hard subset. All rows are \textsf{SKILL.nb} variants, so tokens per success use the same \textsf{SKILL.nb}-internal maintenance and update accounting and are reported in thousands. SR is reported as percent with 95\% Wilson confidence intervals. Fallback is the percent of tasks using natural-language or raw-intent fallback; repair is the percent of tasks with an accepted repair or repository update; regression is update-induced failure after accepted repair or update.}\label{tab:component_ablation}
	\begin{tabular}{@{}lccccc@{}}
		\toprule
		Variant & SR (95\% CI) & Tok. / succ. & Fallback & Repair & Regression \\
		\midrule
		\textsf{SKILL.nb} & 38.4 [32.6, 44.4] & 36k & 18.0\% & 73.0\% & 3.3\% \\
		\(\text{\textsf{SKILL.nb}}_{\text{NL-only}}\) & 33.3 [27.8, 39.3] & 57k & 34.0\% & 61.0\% & 8.8\% \\
		\(\text{\textsf{SKILL.nb}}_{\text{code-only}}\) & 31.0 [25.7, 36.9] & 42k & -- & 48.0\% & 14.7\% \\
		\(\text{\textsf{SKILL.nb}}_{\text{no-gates}}\) & 32.6 [27.1, 38.5] & 45k & 16.0\% & 66.0\% & 18.6\% \\
		\(\text{\textsf{SKILL.nb}}_{\text{no-demote}}\) & 34.1 [28.6, 40.1] & 47k & 20.0\% & 67.0\% & 12.4\% \\
		\bottomrule
	\end{tabular}
\end{table}

\textsf{SKILL.nb} has the highest point SR and the lowest internal token cost per success. Removing code formalization increases cost and fallback use, removing fallback lowers SR and increases regression, and removing gates or lifecycle cleanup increases regression. Because this is a single-stream online ablation on the hard subset, we interpret the pattern as diagnostic evidence that the components are useful rather than as a benchmark-wide causal estimate.

\subsection{Adaptive Threshold Ablation}\label{app:threshold-ablation}

\begin{figure}[t]
	\centering
	\includegraphics[width=\linewidth]{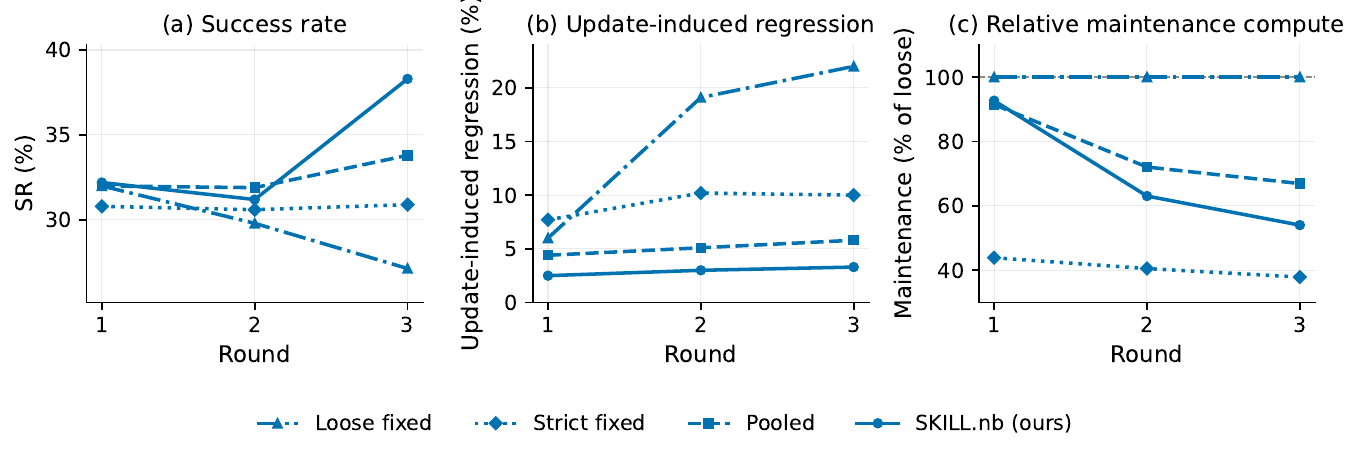}
	\caption{Diagnostic adaptive-threshold ablation on the WebArena-Verified hard subset (258 tasks) over three rounds.
		The four \textsf{SKILL.nb} variants differ only in how lifecycle thresholds are chosen: loose fixed thresholds, strict fixed thresholds, pooled thresholds from the offline estimator without group specialization, and the deployed group-specialized policy (\textsf{SKILL.nb}, ours).
		(a)~Success rate by round.
		(b)~Update-induced regression on previously passing cached traces, by round.
		(c)~Cumulative maintenance compute, normalized to the loose fixed variant at the same round (loose fixed $= 100\%$).
	}\label{fig:threshold_ablation}
\end{figure}

In this section, we include a diagnostic ablation to isolate the contribution of the learned lifecycle policy from the workflow representation itself. Figure~\ref{fig:threshold_ablation} reports this ablation on the 258-task WebArena-Verified hard subset over three rounds. It compares four \textsf{SKILL.nb} variants: the loose fixed variant uses one permissive fixed threshold vector shared across all groups and rounds, the strict fixed variant uses a more restrictive fixed vector, the pooled variant uses pooled thresholds from the offline replay estimator without group specialization, and \textsf{SKILL.nb} (ours) uses the full deployed group-specialized policy. In the current setup, the loose fixed vector is $(\tau_{\mathrm{create}}, \tau_{\mathrm{form}}, \tau_{\mathrm{demote}}, \tau_{\mathrm{retire}}) = (4, 3, 5, 0.50)$ and the strict fixed vector is $(8, 5, 2, 0.25)$. Panel~(a) asks whether specialization improves task success, panel~(b) whether ordinary lifecycle promotions introduce regressions on previously passing cached traces, and panel~(c) whether those gains come with lower cumulative maintenance compute relative to the loose fixed baseline.

Read together, the three panels are consistent with the group-specialized lifecycle policy improving the success--regression--maintenance trade-off on this subset. By round~3, \textsf{SKILL.nb} reaches the highest success rate (38.3\%) while maintaining the lowest update-induced regression (3.3\%), and panel~(c) shows that its cumulative maintenance compute remains below the loose fixed baseline. The fixed-threshold variants illustrate the failure modes avoided. The loose fixed variant over-promotes, driving success down from 32.0\% to 27.1\% while regression rises to 22.0\%. The strict fixed variant under-promotes, producing stable but stagnant success near 31\%. The pooled variant improves on both fixed policies but remains worse than \textsf{SKILL.nb} on all three metrics. The strict fixed variant's low maintenance footprint in panel~(c) reflects this low update activity rather than better efficiency per successful task. The brief round-2 dip for \textsf{SKILL.nb} mirrors the lifecycle result and is followed by the strongest round-3 recovery.

\end{document}